\documentclass[11pt]{article}
\usepackage{graphicx}
\usepackage{subcaption} 
\usepackage{threeparttable}

\usepackage[preprint]{acl}
\usepackage{fontawesome5}
\usepackage{hyperref} 
\hypersetup{
    colorlinks=true,
    urlcolor=blue,
}
\usepackage{times}
\usepackage{latexsym}
\usepackage{fixltx2e}
\usepackage[T1]{fontenc}

\usepackage[utf8]{inputenc}

\usepackage{microtype}
\usepackage{amsmath}
\usepackage{inconsolata}

\usepackage{graphicx}

\usepackage{latexsym}

\usepackage[T1]{fontenc}

\usepackage[utf8]{inputenc}

\usepackage{microtype}

\usepackage{inconsolata}

\usepackage[table]{xcolor}
\usepackage{wrapfig}
\usepackage{graphicx}
\usepackage{amssymb}
\usepackage{amsmath}
\usepackage{subcaption}
\usepackage{subcaption}
\usepackage{booktabs}
\usepackage{multirow}
\usepackage{adjustbox}
\usepackage{colortbl}
\usepackage{subcaption}
\usepackage{placeins}
\usepackage{float}
\usepackage{dblfloatfix}
\usepackage{breqn}
\usepackage{amsthm}
\usepackage{amsmath,amssymb,amsthm}
\usepackage{makecell}
\theoremstyle{plain}

\theoremstyle{definition}

\theoremstyle{remark}

\title{
  \raisebox{-0.4em}{\includegraphics[height=1.5em]{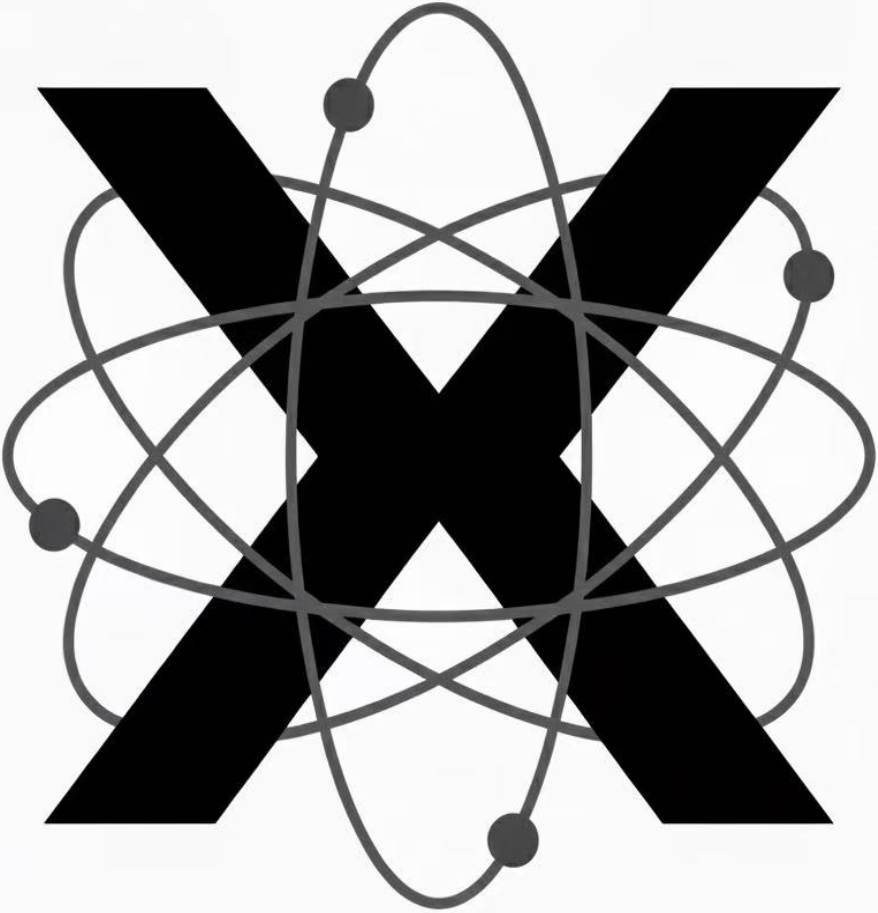}}\hspace{0.5em}
  MMFormalizer: Multimodal Autoformalization \textit{in the Wild}
}

\author{
    \textbf{Jing Xiong}\textsuperscript{1}, 
    \textbf{Qi Han}\textsuperscript{1}, 
    \textbf{Yunta Hsieh}\textsuperscript{2}, 
    \textbf{Hui Shen}\textsuperscript{2}, 
    \textbf{Huajian Xin}\textsuperscript{3}, \\
    \textbf{Chaofan Tao}\textsuperscript{1}, 
    \textbf{Chenyang Zhao}\textsuperscript{6},
    \textbf{Hengyuan Zhang}\textsuperscript{1}, 
    \textbf{Taiqiang Wu}\textsuperscript{1}, 
    \textbf{Zhen Zhang}\textsuperscript{4}, \\
    \textbf{Haochen Wang}\textsuperscript{1}, 
    \textbf{Zhongwei Wan}\textsuperscript{5}, 
    \textbf{Lingpeng Kong}\textsuperscript{1}, 
    \textbf{Ngai Wong}\textsuperscript{1} \\
    \\
    \textsuperscript{1}The University of Hong Kong \quad
    \textsuperscript{2}University of Michigan, Ann Arbor \quad 
    \textsuperscript{3}University of Edinburgh \\
    \textsuperscript{4}University of California, Santa Barbara \quad
    \textsuperscript{5}Ohio State University \quad 
    \textsuperscript{6}University of California, Los Angeles \\
    \\
    \texttt{junexiong@connect.hku.hk}  \\
    \vspace{0.8em} 
    { \ttfamily \small 
         \href{https://MMFormalizer.github.io/}{%
         \raisebox{-0.35em}{\includegraphics[height=1.5em]{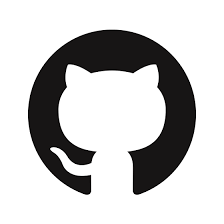}}\
         https://MMFormalizer.github.io/%
         } 
        \quad 
        \href{https://huggingface.co/datasets/menik1126/PhyX-AF}{%
            \raisebox{-0.2em}{\includegraphics[height=1em]{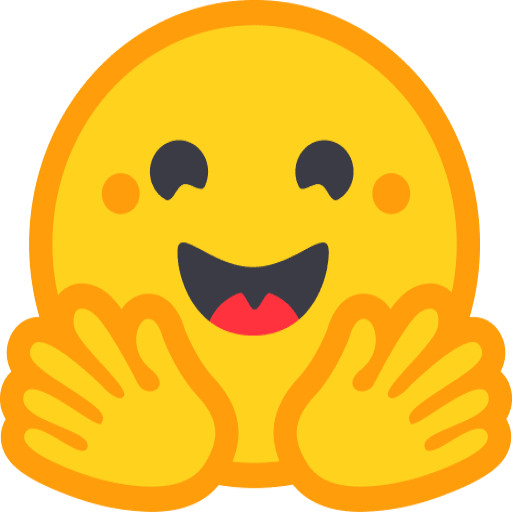}}\ https://huggingface.co/datasets/menik1126/PhyX-AF%
        }
    }
}

\begin{document}
\maketitle
\begin{abstract}

Autoformalization, which translates natural language mathematics into formal statements to enable machine reasoning, faces fundamental challenges \emph{in the wild} due to the multimodal nature of the physical world, where physics requires inferring hidden constraints (e.g., mass or energy) from visual elements. To address this, we propose \textsc{MMFormalizer}, which extends autoformalization beyond text by integrating adaptive grounding with entities from real-world mathematical and physical domains. \textsc{MMFormalizer} \textbf{recursively} constructs formal propositions from perceptually grounded primitives through \emph{recursive grounding} and \emph{axiom composition}, with adaptive recursive \emph{termination} ensuring that every abstraction is supported by visual evidence and anchored in dimensional or axiomatic grounding. We evaluate \textsc{MMFormalizer} on a new benchmark, \textsc{\textbf{PhyX-AF}}, comprising 115 curated samples from \emph{MathVerse}, \emph{PhyX}, \emph{Synthetic Geometry}, and \emph{Analytic Geometry}, covering diverse multimodal autoformalization tasks. Results show that frontier models such as GPT-5 and Gemini-3-Pro achieve the highest compile and semantic accuracy, with GPT-5 excelling in physical reasoning, while geometry remains the most challenging domain. Overall, \textsc{MMFormalizer} provides a scalable framework for unified multimodal autoformalization, bridging perception and formal reasoning. To the best of our knowledge, this is the first multimodal autoformalization method capable of handling classical mechanics (derived from the Hamiltonian), as well as relativity, quantum mechanics, and thermodynamics.


\end{abstract}
\section{Introduction}

Recent advances in large language models (LLMs) show strong capabilities in formal reasoning~\citep{wang2023dt,wang2023lego,xiong2023trigo,xuejun2025mathesis}.
In geometry domain, symbolic reasoning has become a key research frontier~\citep{lu2021inter,murphy2024autoformalizing,zhang2024fgeo,ping2025autogps,he2025matp}, supported by rich geometric data and domain-specific formal languages such as context-free grammar-based predicate forms~\citep{lu2021inter,ping2025autogps} and the Condition Declaration Language (CDL)~\citep{zhang2025diagram,zhang2024fgeo}. However, the above systems rely mainly on symbolic inputs, leaving a gap between visual geometric understanding and formal reasoning, motivating the need for \emph{multimodal autoformalization}. Meanwhile, \texttt{LEAN}~\citep{mathlib4,moura2021lean4} enables rigorous encoding and verification of geometric reasoning, providing a potential pathway.

Multimodal Autoformalization in \emph{geometry}~\citep{murphy2024autoformalizing, he2025matp} emerges as a promising research direction, going beyond the mere recognition of geometric entities in text. This progress is driven not only by the growing interest in connecting perceptual and formal reasoning but also by the extensive collection of geometry-related lemmas and tactics in \texttt{mathlib}~\citep{mathlib4}, which provides a robust foundation for integrating perceptual representations with formal reasoning. Nevertheless, \textit{multimodal autoformalization} beyond the geometric domain remains largely underexplored, particularly in modeling complex phenomena that arise in the real physical world. This is partly due to the lack of supporting infrastructure; although dependencies like PhysLean~\citep{tooby2024formalization} exist, there is still a shortage of tools to integrate them into autoformalization frameworks. To address this issue, we introduce a toolkit for deploying physical theorem search engines and their dependencies.

Another core challenge is: \emph{How can a multimodal autoformalization system be grounded in the wild?} A representative example of this difficulty is Newton’s laws of motion. Although derived from empirical observation and logical reasoning, their formulation reveals a fundamental limitation: relations among quantities such as mass, length, and time cannot be obtained from logic alone but require empirical grounding through measurement. This dependence shows that even a mathematically coherent system must refer back to physical experience through explicit dimensional definitions. Consequently, an autoformalization system cannot arise purely from observation data; it must integrate dimensional analysis as a constraint that links formal statements with empirical interpretability. While nondimensionalization~\citep{buckingham1914physically} abstracts away physical units and offers a potential means to avoid dealing with dimensions, it requires human-provided physical expertise. In contrast, we adopt a dimensional formalism here to maintain a direct bridge between formal reasoning and measurable reality, thereby eliminating the need for intervention by domain experts.

\begin{figure*}[t]
  \centering
  \includegraphics[width=\textwidth]{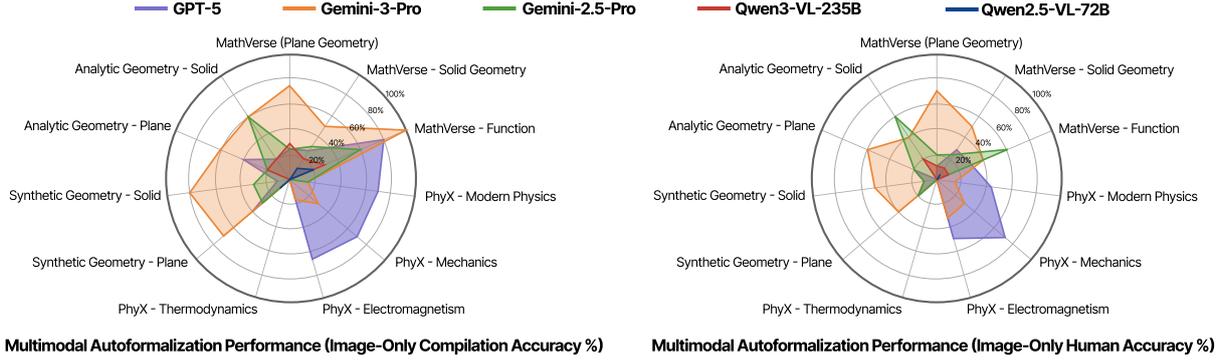}
  \caption{Multimodal autoformalization performance of five representative models across mathematical and physical domains, reported in terms of compilation accuracy (left) and human verification accuracy (right).}
  \label{fig:radar}
\end{figure*}

\begin{figure}[t]
  \centering
  \includegraphics[width=1.0\linewidth]{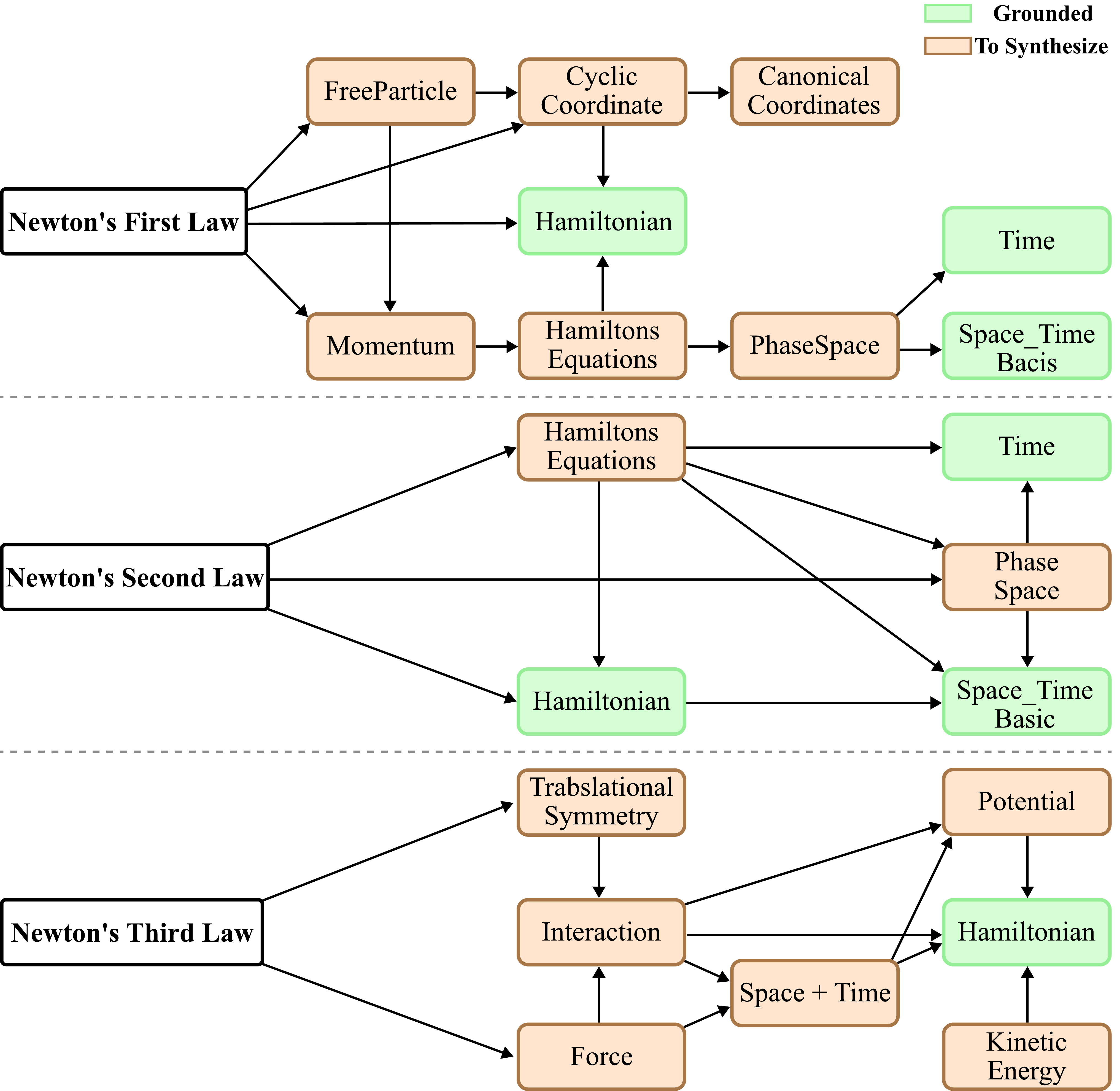}
  \caption{Conceptual dependency graph illustrating how \textit{Newton’s three laws} give rise to key structures in classical mechanics, including momentum, Hamiltonian formulation, phase space, and spacetime representations.}
  \label{fig:newton123}
\end{figure}

A more advanced example is provided by the \textit{theory of relativity}. Rather than being constructed inductively from empirical data, relativity arises from a compact set of foundational axioms—chiefly, the invariance of the speed of light and the equivalence of physical laws across all inertial reference frames. From these axioms, the entire theoretical edifice can be deduced through \textit{thought experiments}, yielding results such as time dilation, length contraction, and the mass–energy equivalence ($E = mc^2$).

Crucially, the above example demonstrates that \textit{identifying the fundamental axioms reveals the most elementary dimensional groundings of the theory}: the unification of space and time through the constant $c$, and the equivalence of mass and energy through the same invariant. In this sense, the \textit{formal system} itself determines what counts as the basic dimensional primitives. We extend a similar idea to the framework of classical mechanics, as illustrated in Fig.~\ref{fig:newton123}, where the three Newtonian laws are derived from the Hamiltonian formalism.

  In this work, we focus on addressing the core challenge of \textit{identifying fundamental axioms} that ground multimodal autoformalization in real-world settings. Our key insight is to recursively ground visual elements through the process of \textit{dimensional grounding} to determine an appropriate point of recursive termination, using the notion of dimension as a bridge between \textit{physical semantics} and \textit{formal logic}. We introduce \textsc{MMFormalizer}, which recursively translates perceptual inputs such as diagrams, text, or physical scenes into structured formal statements by identifying intermediate lemmas, aligning them with symbolic predicates and image elements, and refining them into verifiable axioms within the \texttt{LEAN}~\citep{moura2021lean4} system.

To ensure validity and generality, we build \textsc{PhyX-AF}, a \textit{benchmark} for evaluating multimodal autoformalization across diverse topics. The dataset is designed to minimize redundancy between visual and textual modalities, so that the model cannot rely solely on textual information while ignoring visual input thereby enforcing genuine multimodal reasoning. Our main contributions are threefold:

\begin{itemize}
\item We propose \textsc{MMFormalizer}, a multimodal autoformalization framework that recursively decomposes physical objects into axiomatic steps, translating perceptual elements into consistent formal statements.

\item We introduce tools for searching and interacting with PhysLean, allowing the integration of visual elements into reusable lemmas in \texttt{LEAN}, thereby linking perception with verifiable proofs.

\item We present \textsc{PhyX-AF}, a benchmark for evaluating multimodal autoformalization \emph{in the wild}, which introduces new challenges for assessing formal reasoning.
\end{itemize}

\section{Related Work}
\subsection{Autoformalization}

Autoformalization~\citep{wu2022autoformalization} refers to the process of translating informal mathematical text into formal languages such as Isabelle/HOL~\citep{nipkow2002isabelle}, LEAN~\citep{moura2021lean4}, MetaMath~\citep{megill2019metamath}, and Coq~\citep{barras1999coq}. Previous advancements in this field have expanded the scope to multilingual autoformalization~\citep{jiang2023multilingual}. Later works propose frameworks for semantic alignment, e.g., FormalAlign~\citep{lu2024formalalign}, and retrieval-based methods like RAutoformalizer~\citep{liu2025rethinking}. Recent efforts, such as Kimina-prover~\citep{wang2025kimina} and Mathesis~\citep{xuejun2025mathesis}, emphasize long-horizon formal reasoning to enhance the robustness of the formalization process. Despite notable progress, \emph{multimodal autoformalization} remains an unexplored area.

In this context, \citet{murphy2024autoformalizing} pioneer autoformalization in \texttt{geometry} using \texttt{LEAN}, formalizing multimodal content in Euclidean geometry. However, models still face difficulties with complex geometric figures, such as regular hexagons, which require recursively composing intricate angle relations. To formalize such shapes, the formal system needs to define dependent types that encode geometric constraints, such as equal-length sides and angle constructions, within the \texttt{LEAN}. These constraints are expressed as types parameterized by the underlying geometric properties.

\subsection{Multimodal Formalization}
\textit{Multimodal formalization} integrates symbolic reasoning, formal languages, and geometric problem solving to build interpretable reasoning systems. \citet{lu2021inter} address geometry problems by transforming them into formal symbolic representations grounded in first-order predicate logic. Similarly, \citet{ping2025autogps} propose an automated geometry problem-solving framework that deductive reasoning within a first-order logical language. \citet{zhang2024fgeo} introduce a neuro-symbolic system that combines a many-sorted first-order logic–based formal language with a hypergraph neural network to achieve readable and traceable human-like geometric reasoning. \citet{zhang2025diagram} present DFE-GPS, a multimodal geometry solver that leverages a many-sorted first-order logic formal language to improve diagram comprehension and reasoning.

While first-order and many-sorted logical formal languages provide a structured and interpretable foundation for symbolic geometric reasoning, they fundamentally differ from type-theoretic frameworks such as the dependent type theory underlying \texttt{LEAN}. In first-order logic, entities are abstract symbols belonging to predefined sorts (e.g., \textit{Point}, \textit{Line}, \textit{Circle}), and reasoning proceeds through predicate inference over fixed domains. By contrast, dependent type theory treats mathematical objects as constructive types, where the existence and properties of an object depend on previously defined data. This paradigm enables not only logical deduction but also object construction—turning “proofs” into verifiable computational objects.

The connection between these paradigms lies in their shared goal of achieving rigor and interpretability: first-order logical formalization captures relational structure and deductive closure, while dependent type theory extends this framework into a constructive, computational foundation. Consequently, while systems such as FormalGeo~\citep{zhang2023formalgeo} excel at symbolic geometric reasoning within a fixed formal universe, \textsc{Lean}-style frameworks can represent and manipulate mathematical objects that depend on parameters, constraints, and even continuous quantities.

This distinction becomes crucial for \textit{multimodal autoformalization in the wild}, where models must describe and reason about complex physical entities with \emph{dimensions} rather than purely symbolic ones. Let us continue with the example in Figure~\ref{fig:newton123}. Although we can mathematically and rigorously derive the classical Newtonian mechanical system from the Hamiltonian, it still depends on the definition of its dimensional quantity—namely, energy. Such dependency cannot be adequately captured within a first-order or many-sorted logical formal language, where symbols and relations are dimensionless abstractions. By contrast, in dependent type theory, the dimensional structure of quantities (e.g., length, time, mass, energy) can be encoded directly into the type system, ensuring that all physical expressions are well-typed and dimensionally consistent by construction. Thus, entities like the Hamiltonian, which intrinsically carry physical dimensions, can only be faithfully and constructively defined within a type-theoretic framework.



\begin{figure*}[t]
  \centering
  \includegraphics[width=1.0\linewidth]{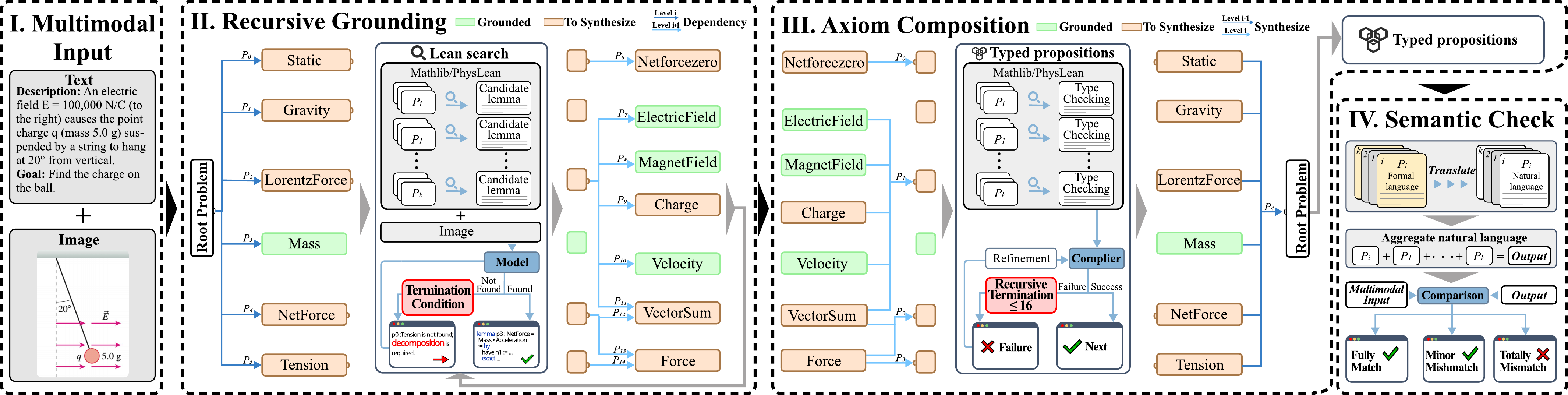}
  \vspace{-4mm}
  \caption{The pipeline overview consists of three stages:
\textit{Recursive Grounding}, identifying physical primitives (the \emph{red} parts in the figure, e.g., the Hamiltonian or dimensional quantities) for \textit{Termination}, and \textit{Axiom Composition}. The \emph{blue} parts in the figure indicate the compiler checking process. The \emph{green} part indicates the formal statements we retrieved from the dependency library.}
  \vspace{-4mm}
  \label{fig:position_id3}
\end{figure*}

\section{Multimodal Autoformalization}

In this section, we demonstrate how complex physical entities can be synthesized from a compact set of primitive constructors. 
Higher-order formal structures (e.g., $\mathsf{PropChain}$, $\mathsf{SceneGraph}$) are recursively composed from perceptually grounded primitives and visual evidence, forming a hierarchy of formally consistent representations.

\subsection{Recursive Grounding}

Given an input image $I : \mathsf{Image}$, we define a \emph{recursive grounding process}:
\begin{equation}
\mathsf{RG} : I \to \mathsf{PropChain},
\end{equation}
which incrementally constructs a dependent hierarchy of formal propositions whose semantics are grounded in perceptual data. 
This process is operationalized by the $\textsc{MMFormalizer}$ model in three inductive stages.

\paragraph{Definition of Lemma.}
We define a \emph{lemma} as a dependent pair consisting of a formal proposition and its constructive inhabitant:
\begin{equation}
\mathsf{Lemma} := \Sigma (P : \mathsf{PropChain}),\; (p : P),
\end{equation}
where $P$ denotes a formal statement, and $p : P$ represents its proof term within the underlying type-theoretic system.

\paragraph{Propositional Grounding.}
A \emph{propositional grounding} is formalized as a dependent chain of lemmas:
\begin{equation}
\mathsf{PropChain} := \Sigma (L_t : \mathsf{List\;Lemma}),
\end{equation}
where each element $L_t$ corresponds to a perceptually grounded proposition indexed by $t$, denoting its position or level within the recursive grounding hierarchy. 
This chain ensures formal consistency across layers while preserving compositional dependencies between propositions.

Formally, the recursive grounding establishes a compositional mapping between perceptual scene structures and propositional representations:
\begin{equation}
\mathsf{lift} : \mathsf{SceneGraph} \to \mathsf{PropChain},
\end{equation}
such that for each subgraph $G_t \subseteq \mathsf{SceneGraph}$, there exists a corresponding lemma $L_t$ within the propositional chain.

\paragraph{Visual Decomposition.}
Let $\mathsf{SceneGraph}$ denote a dependent type encoding primitive visual entities and their spatial relations:
\begin{equation}
\mathsf{SceneGraph} := \Sigma (V_t : \mathsf{List\;Primitive}),\; \mathsf{Rel}(V_t),
\end{equation}
where $V_t = [v_1, v_2, \dots, v_n]$ is a list of \emph{visual primitives}, each $v_i : \mathsf{Primitive}$ representing a perceptual entity such as a point, line, or region. The type $\mathsf{Primitive}$ thus enumerates the basic perceptual elements, formally defined as
\begin{equation}
\mathsf{Primitive} ::= \texttt{point} \mid \texttt{line} \mid \texttt{region}.
\end{equation}
The dependent component $\mathsf{Rel}(V_t)$ specifies the spatial or topological relations among the primitives in $V_t$,
\begin{equation}
\mathsf{Rel}(V_t) : \Pi (v_i, v_j \in V_t),\; \mathsf{SpatialRel}(v_i, v_j),
\end{equation}
where $\mathsf{SpatialRel}$ captures geometric or structural predicates such as \texttt{adjacent}, \texttt{parallel}, or \texttt{contained\_in}. The parsing function
\begin{equation}
\mathsf{parse} : I \to \mathsf{SceneGraph}
\end{equation}
decomposes an input image $I$ into a base-level scene graph $G_0 : \mathsf{SceneGraph}$.

Each primitive $v_i \in V_0$ within $G_0$ is assigned an initial semantic hypothesis $l_t$, which is typically a short phrase as illustrated by the \textcolor{brown}{brown} regions in Figure~\ref{fig:newton123}. These hypotheses constitute the perceptual priors of the \emph{grounding} chain $\mathsf{RG}(I)$, forming the base layer $L_0$ within the propositional structure.

\subsection{Recursive Termination}

We define a mapping
\begin{equation}
\mathsf{map} : ( V_t, l_t) \to L_t,
\end{equation}
which maps each visual element $V_t$ into the space of axioms,
thereby constructing a hierarchy of verifiable axiomatic statements within the \texttt{LEAN}. At each stage, we ensure that every formal abstraction is both supported by perceptual grounding
and capable of constraining subsequent semantic parsing. We implement the above process through two major phases: the \textit{Grounding} and the \textit{Termination}, culminating in termination through axiomatic or dimensional grounding.

\paragraph{Grounding.}
We use the decomposed intermediate lemma statements as queries to search for theorems in \texttt{LeanSearch}~\citep{gao2024semantic}, retrieving them from \texttt{mathlib}~\citep{mathlib4} and \texttt{PhysLean}~\citep{tooby2024formalization}.

In local deployment, \texttt{LeanSearch} runs on a \texttt{LEAN} + \texttt{mathlib} + \texttt{PhysLean} environment, uses indexing scripts to extract all declarations (theorems, lemmas, definitions, structures) together with their type signatures and comments, encodes these statements into embeddings, and enables offline semantic search to retrieve the most top-$k$ relevant formal items for any $L_t$. For each node $t \in \mathbb{N}$, we assume a substructure $G_t \subseteq G_{t-1}$ associated with a predicate set $P_t$. Formally, we define
\begin{equation}
P_t := { p(v_i, l_t\mid G_{t-1}) \mid v_i \in V_t ,},
\end{equation}
where $V_t$ denotes the set of visual elements extracted from picture, and $l_t$ represents the predicted informal term that encodes their semantic relationships. The set $P_t$ is thus regarded as a collection of \textit{informal statements}, each expressing a perceptually grounded yet not fully formalized correspondence between the visual configuration and its prospective symbolic abstraction. In this way, $P_t$ serves as a conceptual bridge linking perceptual structures to their formal statements. The alignment operator
\begin{equation}
\mathsf{Grounding} : G_{t-1} \to P_t \to \mathsf{Lemma}
\end{equation}
then maps $p(v_i, l_t)$ into the corresponding formal statements. We then prompt the LLMs to select from these statements the ones that best align with the current $P_t$, and designate it as the $L_t$ This grounding is inductively extended by defining
\begin{equation}
L_{t+1} := \mathsf{Grounding}(G_{t}, P_{t+1}).
\end{equation}

\paragraph{Termination.}
The recursion terminates when $P_t$  is grounded either in a primitive dimensional or an axiom:

\begin{equation}
\begin{aligned}
&\mathsf{Termination}(P_t)=
\begin{cases}
\mathsf{dim}(p), & \text{if } p \in D_t,\\[4pt]
\mathsf{axiom}(p), & \text{if } p \in A_t.
\end{cases}
\end{aligned}
\end{equation}

where $D_t$ denotes the physical dimensional quantity, which in geometric problems usually corresponds to length; $A_t$ denotes the fundamental $\mathsf{axiom}$; $\mathsf{dim}$ maps a physical quantity to its fundamental dimensional basis.

\subsection{Axiom Composition}

At this stage, we define the $\mathsf{Axiom Chain}$ as the formal closure of the propositional hierarchy:
\begin{equation}
\scalebox{0.94}{ $
\mathsf{AxiomChain} := \Sigma (A_t : \mathsf{List\;Axiom}, D_t :\mathsf{List\;Dim})). \;  $ }
\end{equation}
Formally, there exists a grounding mapping
\begin{equation}
\mathsf{ground} : \mathsf{PropChain} \to \mathsf{AxiomChain}.
\end{equation}$\mathsf{AxiomChain}$ represents the final, ontologically closed layer of the multimodal formalization hierarchy, 
where every perceptually grounded proposition is instantiated within a physically or axiomatically formal statements.

Through successive applications of $\mathsf{Compose}$, performed by the LLM, all child nodes are recursively combined to construct a non-leaf node:
\begin{equation}
\mathsf{Compose}({L_{t+1}^k}, G_t, P_t) \to L_t,
\end{equation}
where ${L_{t+1}^k}$ denotes the set of lemmas from the child nodes of node $t$ and $k$ is the number of childs. This process captures the transition from perceptual substructures to hierarchically dependent propositions within the recursive grounding framework. After each composition at node $t$, the resulting formal statements are passed through a syntax checker for compilation verification, ensuring their structural and logical validity within the \texttt{LEAN}.

Each composed type thus preserves both dimensional structure and symbolic manipulability, enabling reasoning over geometric, physical, and logical domains within a unified formal statements. 
In this sense, $\mathsf{TypeComposition}$ acts as the meta-level operator that aligns \texttt{LEAN}’s dependent type hierarchy with the \emph{multimodal autoformalization} pipeline:
\begin{align}
\mathsf{TypeComposition} :\;& \mathsf{SceneGraph} 
  \to \mathsf{PropChain} \nonumber \\[4pt]
  & \to \mathsf{AxiomChain},
\end{align} 
ensuring that every physical entity corresponds to a grounded formal statement. Through such recursive compositionality, \texttt{LEAN} supports the construction of formal models that faithfully reflect the generative structure of physical and conceptual reality—where primitives (points, vectors, forces) combine to form law-constrained, formal systems.

\subsection{Semantic Checking}

The \textit{semantic checking} module verifies whether each formal statement is jointly supported by visual and textual evidence. 
For each image–text–formula triplet, the checker outputs an indicator value, 
where $1$ denotes semantic acceptance and $0$ denotes rejection. 
If all propositions $L_t \in \mathsf{PropChain}$ are assigned $1$, 
the chain is considered valid and grounded across modalities.

\begin{table*}[t]
\centering
\caption{Comparison across \textsc{MathVerse}, \textsc{PhyX}, \textsc{Synthetic Geometry}, and \textsc{Analytic Geometry} datasets. We report Compile accuracy, semantic correctness, and human verification results. The \emph{Modern} category under \textsc{PhyX} includes problems from both quantum mechanics and relativity.}
\vspace{-2mm}
\label{tab:detailed_results_filtered}
\resizebox{\linewidth}{!}{
\begin{tabular}{
@{}lc|
>{\centering\arraybackslash}p{0.6cm} >{\centering\arraybackslash}p{0.6cm} 
>{\centering\arraybackslash}p{0.6cm} >{\centering\arraybackslash}p{0.6cm} 
>{\centering\arraybackslash}p{0.6cm} >{\centering\arraybackslash}p{0.6cm}|
>{\centering\arraybackslash}p{0.6cm} >{\centering\arraybackslash}p{0.6cm} 
>{\centering\arraybackslash}p{0.6cm} >{\centering\arraybackslash}p{0.6cm} 
>{\centering\arraybackslash}p{0.6cm} >{\centering\arraybackslash}p{0.6cm} 
>{\centering\arraybackslash}p{0.6cm} >{\centering\arraybackslash}p{0.6cm}|
>{\centering\arraybackslash}p{0.6cm} >{\centering\arraybackslash}p{0.6cm} 
>{\centering\arraybackslash}p{0.6cm} >{\centering\arraybackslash}p{0.6cm}|
>{\centering\arraybackslash}p{0.6cm} >{\centering\arraybackslash}p{0.6cm}
>{\centering\arraybackslash}p{0.6cm} >{\centering\arraybackslash}p{0.6cm}
@{}}
\toprule
\multicolumn{1}{l|}{\textbf{Model}} &
\multicolumn{1}{c|}{\textbf{Metric}} &
\multicolumn{6}{c|}{\textbf{\textsc{MathVerse}}} &
\multicolumn{8}{c|}{\textbf{\textsc{PhyX}}} &
\multicolumn{4}{c|}{\textbf{\textsc{Synthetic Geometry}}}&
\multicolumn{4}{c}{\textbf{\textsc{Analytic Geometry}}}
\\ 

\multicolumn{1}{l|}{}& &
\multicolumn{2}{c|}{\textbf{\makecell{Plane Geometry}}}&\multicolumn{2}{c|}{\textbf{\makecell{Solid Geometry}}}&\multicolumn{2}{c|}{\textbf{Function}}&
\multicolumn{2}{c|}{\textbf{Modern}}&\multicolumn{2}{c|}{\textbf{Mechanics}}&\multicolumn{2}{c|}{\textbf{\makecell{Electro-\\magnetism}}}&\multicolumn{2}{c|}{\textbf{\makecell{Thero-\\dynamics}}}&
\multicolumn{2}{c|}{\textbf{\makecell{Plane Geometry}}}&\multicolumn{2}{c|}{\textbf{\makecell{Solid Geometry}}}&\multicolumn{2}{c|}{\textbf{Plane Geometry}}&\multicolumn{2}{c}{\textbf{Solid Geometry}}
\\

\multicolumn{1}{l|}{} & &
\multicolumn{1}{c}{\textbf{Img}} & \multicolumn{1}{c|}{\textbf{Text}} &
\multicolumn{1}{c}{\textbf{Img}} & \multicolumn{1}{c|}{\textbf{Text}} &
\multicolumn{1}{c}{\textbf{Img}} & \multicolumn{1}{c|}{\textbf{Text}}  &
\multicolumn{1}{c}{\textbf{Img}} & \multicolumn{1}{c|}{\textbf{Text}} &
\multicolumn{1}{c}{\textbf{Img}} & \multicolumn{1}{c|}{\textbf{Text}}  &
\multicolumn{1}{c}{\textbf{Img}} & \multicolumn{1}{c|}{\textbf{Text}}  &
\multicolumn{1}{c}{\textbf{Img}} & \multicolumn{1}{c|}{\textbf{Text}}  &
\multicolumn{1}{c}{\textbf{Img}} & \multicolumn{1}{c|}{\textbf{Text}}  &
\multicolumn{1}{c}{\textbf{Img}} & \multicolumn{1}{c|}{\textbf{Text}}  &
\multicolumn{1}{c}{\textbf{Img}} & \multicolumn{1}{c|}{\textbf{Text}}  &
\multicolumn{1}{c}{\textbf{Img}} & \multicolumn{1}{c}{\textbf{Text}} 

\\ 
\midrule
\multicolumn{24}{c}{\cellcolor[HTML]{E5E5FC}\textit{\textbf{Frontier Model}}} \\ 
\midrule

\multirow{3}{*}{GPT-5} &
Compile &
24.0 & 8.0 &
28.0 & 8.0 &
80.0 & 0.0 &
71.4 & 37.5 &
71.4 & 16.7 &
66.7 & 50.0 &
0.0 & 20.0 &
40.0 & 30.0 &
10.0 & 20.0 &
40.0 & 0.0 &
20.0 & 100.0 
\\ 
\multicolumn{1}{l}{} &
Semantics &
20.0 & 0.0 &
28.0 & 8.0 &
30.0 & 0.0&
71.4 & 12.5 &
71.4 & 0.0 &
50.0 & 0.0 &
0.0 & 0.0 &
40.0 & 30.0 &
0.0 & 10.0 &
20.0 & 0.0 &
20.0 & 100.0 \\ 
\multicolumn{1}{l}{} &
Human Check &
12.0& -- &
28.0& -- &
30.0 & -- &
42.9 & --&
71.4 & -- &
50.0 & -- &
0.0 & -- &
20.0 & -- &
10.0 & -- &
20.0 & -- &
0.0 & -- \\ 
\midrule

\multirow{3}{*}{Gemini-3-Pro} &
Compile &
76.0 & 8.0 &
52.0 & 4.0 &
100.0 & 40.0 &
14.3 & 57.1 &
28.6 & 42.9 &
16.7 & 16.7 &
0.0 & 0.0 &
70.0 & 40.0 &
80.0 & 70.0 &
60.0 & 0.0 &
60.0 & 40.0 \\ 
\multicolumn{1}{l}{} &
Semantics &
76.0 & 0.0 &
52.0 & 0.0 &
40.0 & 0.0 &
14.3 & 42.9 &
28.6 & 28.6 &
33.3 & 0.0 &
0.0 & 0.0 &
70.0 & 30.0 &
80.0 & 70.0 &
60.0 & 0.0 &
40.0 & 20.0 \\ 
\multicolumn{1}{l}{} &
Human Check &
72.0 & -- &
52.0 & -- &
40.0 & -- &
14.3& -- &
28.6& -- &
33.3& -- &
0.0 & -- &
40.0 & -- &
50.0 & -- &
60.0 & -- &
40.0& -- \\ 
\midrule

\multirow{3}{*}{Gemini-2.5-Pro} &
Compile &
24.0 & 8.0 &
32.0 & 8.0 &
60.0 & 60.0 &
14.3 & 0.0 &
0.0& 0.0 &
0.0 & 0.0 &
0.0 & 20.0 &
30.0 & 30.0 &
30.0 & 30.0 &
20.0 & 0.0 &
60.0 & 0.0 \\ 
\multicolumn{1}{l}{} &
Semantics &
20.0 & 0.0 &
24.0 & 0.0 &
60.0 & 0.0 &
14.3 & 0.0 &
0.0 & 0.0 &
0.0 & 0.0 &
0.0 & 20.0 &
30.0 & 30.0 &
10.0 & 20.0 &
20.0 & 0.0 &
40.0 & 0.0 \\ 
\multicolumn{1}{l}{} &
Human Check &
20.0 & -- &
24.0 & -- &
60.0& -- &
0.0 & -- &
0.0& -- &
0.0& -- &
0.0& -- &
20.0 & -- &
10.0& -- &
20.0& -- &
60.0& -- \\ 
\bottomrule

\multicolumn{24}{c}{\cellcolor[HTML]{E5E5FC}\textit{\textbf{Open-source Model}}} \\ 
\midrule

\multirow{3}{*}{Qwen3-VL-235B} &
Compile &
28.0 & 4.0 &
20.0 & 0.0 &
30.0 & 30.0 &
0.0 & 0.0 &
0.0 & 0.0 &
0.0 & 0.0 &
0.0 & 0.0 &
0.0 & 10.0 &
0.0 & 0.0 &
20.0 & 0.0 &
20.0 & 0.0 \\ 
\multicolumn{1}{l}{} &
Semantics &
16.0 & 0.0 &
16.0 & 0.0 &
20.0 & 0.0 &
0.0 & 0.0 &
0.0 & 0.0 &
0.0 & 0.0 &
0.0 & 0.0 &
0.0 & 10.0 &
0.0 & 0.0 &
20.0 & 0.0 &
20.0 & 0.0 \\ 
\multicolumn{1}{l}{} &
Human Check &
12.0 & -- &
12.0 & -- &
10.0 & -- &
0.0 & -- &
0.0 & -- &
0.0 & -- &
0.0 & -- &
0.0 & -- &
0.0 & -- &
0.0 & -- &
20.0 & -- \\ 
\midrule

\multirow{3}{*}{Qwen2.5-VL-72B} &
Compile &
0.0 & 0.0 &
12.0 & 0.0 &
20.0 & 20.0 &
0.0 & 0.0 &
0.0 & 0.0 &
0.0 & 0.0 &
0.0 & 0.0 &
10.0 & 0.0 &
0.0 & 0.0 &
0.0 & 0.0 &
0.0 & 0.0 
\\ 
\multicolumn{1}{l}{} &
Semantics &
0.0 & 0.0 &
8.0 & 0.0 &
0.0& 0.0&
0.0 & 0.0 &
0.0 & 0.0 &
0.0 & 0.0 &
0.0 & 0.0 &
0.0 & 0.0 &
0.0 & 0.0 &
0.0 & 0.0 &
0.0 & 0.0 \\ 
\multicolumn{1}{l}{} &
Human Check &
0.0 & -- &
4.0 & -- &
0.0 & --&
0.0 & -- &
0.0 & -- &
0.0 & -- &
0.0 & -- &
0.0 & -- &
0.0& -- &
0.0& -- &
0.0& -- \\ 
\bottomrule

\end{tabular}
}
\label{tab:detailed_results_filtered}
\end{table*}

\begin{table*}[t]
\centering
\renewcommand{\arraystretch}{0.85}
\setlength{\tabcolsep}{6pt}
\caption{Ablation Study on the Model Components: Evaluating the Effect of Different Modifications on Performance. The experiments include: (1) \emph{Ablation on synthesizer without code}: We did not provide the model with reference code while synthesizing new types. (2) \emph{Ablation on termination condition}: We explicitly specified the decomposition recursion termination condition. (3) \emph{Ablation on grounding with image}: We investigated the impact of inputting images during the grounding process. (4) \emph{Ablation on pass@k}: We set sampling attempts (\textit{k=3}) for each node.}
\label{tab:ablation}
\resizebox{\linewidth}{!}{
\begin{tabular}{
@{}c|ccc|cccc|cc|cc
@{}}
\toprule
\multicolumn{1}{c|}{\textbf{Metric}} &
\multicolumn{3}{c|}{\textbf{\textsc{MathVerse}}} &
\multicolumn{4}{c|}{\textbf{\textsc{PhyX}}} &
\multicolumn{2}{c|}{\textbf{\textsc{Synthetic Geometry}}} &
\multicolumn{2}{c}{\textbf{\textsc{Analytic Geometry}}}
\\

 &
\multicolumn{1}{c}{\makecell{Plane \\ Geometry}} &
\multicolumn{1}{c}{\makecell{Solid \\ Geometry}} &
\multicolumn{1}{c|}{Function} &
\multicolumn{1}{c}{Modern} &
\multicolumn{1}{c}{Mechanics} &
\multicolumn{1}{c}{\makecell{Electro- \\ magnetism}} &
\multicolumn{1}{c|}{\makecell{Thero- \\ dynamics}} &
\multicolumn{1}{c}{\makecell{Plane \\ Geometry}} &
\multicolumn{1}{c|}{\makecell{Solid \\ Geometry}} &
\multicolumn{1}{c}{\makecell{Plane \\ Geometry}} &
\multicolumn{1}{c}{\makecell{Solid \\ Geometry}}
\\

\midrule

\rowcolor[HTML]{E5E5FC}
\multicolumn{12}{c}{\textit{\textbf{original}}} \\
\midrule

Compile &
50.0 & 33.3 & 66.7&
50.0 & 60.0 & 100.0 & 100.0 &
33.3 & 0.0 &
33.3 & 0.0
\\

Semantics &
25.0& 33.3 & 66.7 &
50.0 & 60.0 & 100.0 & 0.0 &
33.3 & 0.0 &
0.0 & 0.0
\\

Average nodes &
21.8 & 10.3 & 14.3 &
27.3& 15.0 & 13.5 & 35.0 &
15.7 & 21.3 &
12.7 & 30.0
\\

Average depth &
3.4 & 5.0 & 3.0 &
3.0 & 4.4 & 4.5 & 6.0 &
2.3 & 4.3 &
3.3 & 5.0
\\

\midrule

\rowcolor[HTML]{E5E5FC}
\multicolumn{12}{c}{\textit{\textbf{Ablation on synthesizer without code}}} \\
\midrule

Compile &
25.0 & 33.0 & 100.0 &
100.0 & 40.0 & 0.0 & 100.0 &
33.3 & 66.7 &
0.0 & 0.0
\\

Semantics &
25.0 & 16.7 & 33.3 &
100.0 & 40.0 & 0.0 & 0.0 &
33.3 & 66.7 &
0.0 & 0.0
\\

\midrule

\rowcolor[HTML]{E5E5FC}
\multicolumn{12}{c}{\textit{\textbf{Ablation on termination condition}}} \\
\midrule

Average nodes &
34.3 & 24.3 & 14.3 &
5.5 & 16.4 & 31.5 & 28.0 &
16.0 & 22.0 &
17.0 & 21.0
\\

Average depth &
4.3 & 4.0 & 3.0 &
3.0 & 6.6 & 7.0 & 8.0 &
3.0 & 5.3 &
3.7 & 3.0
\\

\midrule

\rowcolor[HTML]{E5E5FC}
\multicolumn{12}{c}{\textit{\textbf{Ablation on grounding with image}}} \\
\midrule

Compile &
25.0 & 33.0 & 100.0 &
100.0 & 80.0 & 50.0 & 0.0 &
33.3 & 66.7 &
0.0 & 0.0
\\

Semantics &
25.0 & 16.7 & 66.7 &
100.0 & 60.0 & 50.0 & 0.0 &
33.3 & 66.7 &
0.0 & 0.0
\\

\midrule

\rowcolor[HTML]{E5E5FC}
\multicolumn{12}{c}{\textit{\textbf{Ablation on pass@k}}} \\
\midrule

Compile &
50.0 & 66.7 & 100.0 &
100.0 & 80.0 & 0.0 & 0.0 &
66.7 & 100.0 &
10.0 & 0.0
\\

Semantics &
25.0 & 16.7 & 33.3 &
100.0 & 60.0 & 0.0 & 0.0 &
66.7 & 66.7 &
0.0 & 0.0
\\

\bottomrule
\end{tabular}
}
\vspace{-4mm}
\end{table*}

\begin{table}[!htbp]
\centering
\caption{Accuracy of Semantic Checking by Different LLMs, evaluated through human validation.
The table reports the agreement rate between each model’s semantic checking and human verification on \texttt{LEAN} code generated by \textbf{G} (GPT-5) and \textbf{Q} (Qwen3-VL-235B). $\diamond$ indicates that the model failed to generate compilable code, thereby precluding human verification.}
\label{tab:single_column_acc_only}

\small 
\renewcommand{\arraystretch}{1.0} 
\setlength{\tabcolsep}{3pt} 

\begin{tabular*}{\columnwidth}{@{\extracolsep{\fill}} c l c c @{}}
\toprule
\multirow{2}{*}{\textbf{Data}} & \multirow{2}{*}{\textbf{Model}} & \multicolumn{2}{c}{\textbf{Accuracy (\%)}} \\
\cmidrule(l){3-4} 
 & & \textbf{G} & \textbf{Q} \\
\midrule

\multirow{5}{*}{\textbf{\shortstack{\textsc{MathVerse}}}} 
& GPT-5 & 69.2 & 76.5 \\
& Gemini-2.5-Pro & \textbf{84.6} & \textbf{88.9} \\
& Gemini-3-Pro & 84.6 & 77.9\\
& Qwen3-VL-235B & 30.8 & 66.7 \\
& Qwen2.5-VL-72B & 15.4& 77.9 \\

\midrule

\multirow{5}{*}{\textbf{\textsc{PhyX}}} 
& GPT-5 & 71.4 & $\diamond$ \\
& Gemini-2.5-Pro & \textbf{78.6} & $\diamond$ \\
& Gemini-3-Pro & 78.6 & $\diamond$ \\
& Qwen3-VL-235B & 53.9 & $\diamond$  \\
& Qwen2.5-VL-72B & 76.9 & $\diamond$  \\

\midrule

\multirow{5}{*}{\textbf{\scriptsize \textsc{\textsc{Synthetic Geometry}}}} 
& GPT-5 & 40.0 &$\diamond$\\
& Gemini-2.5-Pro &40.0 & $\diamond$ \\
& Gemini-3-Pro & 60.0 & $\diamond$ \\
& Qwen3-VL-235B & 60.0& $\diamond$ \\
& Qwen2.5-VL-72B & \textbf{60.0} & $\diamond$ \\
\midrule

\multirow{5}{*}{\textbf{\scriptsize \textsc{Analytic Geometry}}} 
& GPT-5 & 0.0 & 100.0 \\
& Gemini-2.5-Pro & \textbf{66.7} & 50.0 \\
& Gemini-3-Pro & 33.3 & 100.0 \\
& Qwen3-VL-235B & 33.3 & 100.0 \\
& Qwen2.5-VL-72B &33.3& \textbf{100.0} \\

\bottomrule
\end{tabular*}
\label{tab:Semantic Checking}
\end{table}

\section{Experiment}

\begin{figure}[t]
  \centering
  \includegraphics[width=1.0\linewidth]{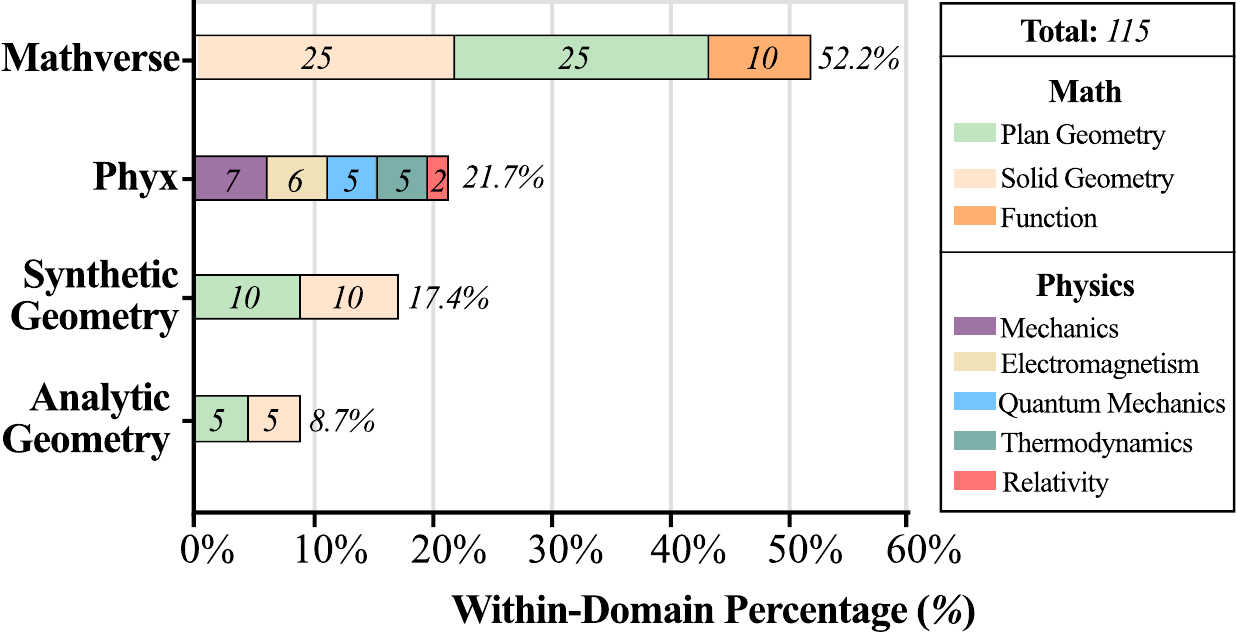}
  \caption{Distribution of Problem Types.}
  \label{fig:position_id3}
\end{figure}
\subsection{Experimental Setup}

\paragraph{PhyX-AF.}
To rigorously evaluate the \emph{multimodal autoformalization} capability, we construct the \textsc{PhyX-AF} benchmark comprising 115 meticulously curated samples drawn from representative sources, including \textsc{MathVerse}~\citep{zhang2024mathverse}, \textsc{PhyX}~\citep{shen2025phyx}, and \textsc{Synthetic Geometry}~\citep{trinh2024solving}, which is further augmented with a rule-based computation engine~\citep{hubert2025olympiad} for automated synthesis and verification, as well as an extended \textsc{Analytic Geometry} dataset designed to assess autoformalization within coordinate-based geometric contexts. We present the dataset statistics in Figure~\ref{fig:position_id3}, which illustrates that the benchmark covers both \emph{mathematics} and \emph{physics} domains.

\paragraph{Data Filtering.}
To ensure genuine multimodal reasoning, we apply a strict \emph{visual dependency criterion}: only problems where the diagram is indispensable for solving are retained; samples solvable by text alone are removed.

\paragraph{Multimodal Mathematical Setup.}
This setup primarily samples data from \textsc{MathVerse}, covering three categories: \textit{Plane Geometry}, \textit{Solid Geometry}, and \textit{Function}, mainly drawn from a \textit{real-exam setting} with authentic geometry problems designed for \textit{in-distribution} autoformalization.
Within this setup, the \textit{Function} category mainly involves solving multivariate and higher-order equations based on given function graphs, while the \textit{Plane Geometry} and \textit{Solid Geometry} categories primarily focus on proof-based problems.

\vspace{-1mm}
\paragraph{Analytic Geometry.}
We include analytic geometry problems extracted from \textsc{Geometry3K}~\citep{lu2021inter} and \textsc{GeoInt}~\citep{wei2025geoint}, as well as procedurally generated 2D and 3D figures composed of geometric primitives (points, lines, planes, arcs, polyhedra). This task requires LLMs to understand both geometric and numerical relations, often involving uncommon or composite numerical and visual configurations that are absent from \texttt{mathlib} or other pretraining corpora.


\paragraph{Multimodal Physical Setup.}
This setup involves visual scenes and physical interactions from natural or simulated environments, assessing \textit{visual–physical generalization}. The \textsc{PhyX} subset (21.7\%) includes problems from mechanics, electromagnetism, thermodynamics, and relativity and quantum mechanics. Each image is paired with a textual statement specifying geometric or physical constraints, evaluating end-to-end \textit{visual-to-formal grounding} under realistic perceptual noise.

\paragraph{Synthetic Geometry Settings.}
In this setup, we aim to test the model’s ability to synthesize new dependent types and constructors at test time, thereby evaluating \textit{out-of-distribution} autoformalization and formal generalization beyond its pretrained knowledge. Many constructed geometric objects and relational schemas \textit{do not exist in \texttt{mathlib}} or other pretrained corpora. Thus, the \textsc{mmformalizer} must create \textit{novel dependent types at test time}, testing its ability to generalize beyond its training distribution. This setup evaluates \textit{out-of-distribution} autoformalization and synthesis of new \textit{type constructors} for unseen perceptual structures. Following AlphaGeometry~\citep{trinh2024solving}, we adopt a symbolic deduction engine for Olympiad-level geometric synthesis. The specific synthesis rules and verification setup can be found in Appendix~\ref{Synthetic Geometry Setup}.

\vspace{-2mm}
\subsection{Main Results}
Our main experimental results are presented in Table~\ref{tab:detailed_results_filtered}. 
Our main findings are as follows: 
(i) \emph{Frontier models demonstrate stronger multimodal reasoning overall.} 
Among all evaluated systems, Gemini-3-Pro achieves the highest overall compile and semantic accuracy on \textsc{MathVerse} and \textsc{Analytic Geometry}, while \textsc{GPT-5} shows a clear advantage on the \textsc{PhyX} dataset. 
In particular, \textsc{GPT-5} performs notably better in the \textsc{Modern} category of \textsc{PhyX}, which includes quantum mechanics and relativity problems, reflecting stronger physical reasoning and grounding. (ii) \emph{Geometry reasoning remains challenging.} All models exhibit significantly lower accuracy on the \textsc{Synthetic Geometry} and \textsc{Analytic Geometry} subsets, indicating persistent difficulties in bridging visual reasoning with formal reasoning. 
It suggests that models still face challenges in accurately understanding concrete length and angle relationships, as well as generalizing to distributions beyond those seen during training. Even advanced LLMs such as Gemini-3-Pro and Gemini-2.5-Pro show large performance gaps between image and text modalities. (iii) \emph{Advanced open-source model lag behind frontier models.} The most powerful open-source model, Qwen3-VL-235B, is almost unable to solve problems in the physical domain or in out-of-distribution synthetic geometry tasks.

\vspace{-1mm}
\subsection{What matters in \textsc{MMFormalizer}?}
In this section, we perform ablation studies to analyze how several design factors affect model performance, including (1) using only retrieved theorem names instead of full code snippets (synthesizer without code), (2) applying an explicit recursion termination condition (termination condition), (3) incorporating images during the grounding stage (grounding with image), and (4) enabling parallel sampling for each node (\emph{pass@k}). As presented in Table~\ref{tab:termination_logic_comparison}, we make the following observations: \textit{(i)} For the \textsc{synthetic geometry} setting where the data are likely never seen in the pre-training corpus we find that removing the retrieved reference code significantly improves model performance. This suggests that allowing the model to synthesize freely, rather than constraining its output space with retrieved reference code, can substantially enhance its out-of-distribution generalization ability. \textit{(ii)} Not specifying a detailed termination condition leads to an excessively deep recursive tree and an overly large dependency graph, which eventually causes the synthesis process to fail. \textit{(iii)} \emph{Grounding with image} can significantly enhance performance in more challenging settings, such as the \emph{Modern Physics} category, which includes relativity and quantum mechanics, as well as in \emph{Synthetic Geometry}. \textit{(iv)} Increasing the sampling number (\emph{pass@k}) can improve performance on more difficult problems, indicating that test-time scaling holds great potential for \textsc{MMFormalization}.

\subsection{Semantic Checking Analysis}
In this section, we employ the most powerful closed-source model, GPT-5, and the most powerful open-source model, Qwen3-VL-235B, to perform multimodal autoformalization. As shown in Table~\ref{tab:Semantic Checking}, we use the following five models for semantic checking: GPT-5, Gemini-2.5-pro, Gemini-3-pro, Qwen3-VL-235B, and Qwen2.5-VL-72B. Finally, human checking is conducted to verify the accuracy of the \emph{semantic checking} results produced by these five models. We have the following main findings: (i) In the \textsc{MathVerse}, \textsc{PhyX}, and \textsc{Analytic Geometry} subsets, we find that Gemini-2.5-Pro achieves the highest accuracy. (ii) The weakest Qwen2.5-VL-72B model performs on par with the state-of-the-art Gemini-3-Pro in \emph{semantic checking} tasks using code generated by the open-source Qwen3-VL-235B model, revealing the potential of weak models supervising strong models. (iii) Qwen2.5-VL-72B demonstrates strong semantic verification capabilities, slightly weaker than those of Qwen3-VL-235B.

\vspace{-2mm}
\section{Conclusion}
\vspace{-2mm}
In this work, we presented \textsc{MMFormalizer}, a unified framework for multimodal autoformalization that bridges perceptual understanding and formal reasoning. By recursively grounding visual and textual inputs into verifiable logical structures, our method enables interpretable formalization across mathematics and physics, including classical mechanics, relativity, quantum mechanics, and thermodynamics. Evaluations on the newly proposed \textsc{PhyX-AF} benchmark demonstrate the effectiveness of our approach and reveal both the promise and limitations of current large multimodal models in formal reasoning.

\clearpage

\bibliography{custom}
\clearpage
\appendix

\setcounter{figure}{0}
\renewcommand{\thefigure}{A.\arabic{figure}}

\begin{figure}[!ht]
    \centering
  \includegraphics[width=\linewidth]{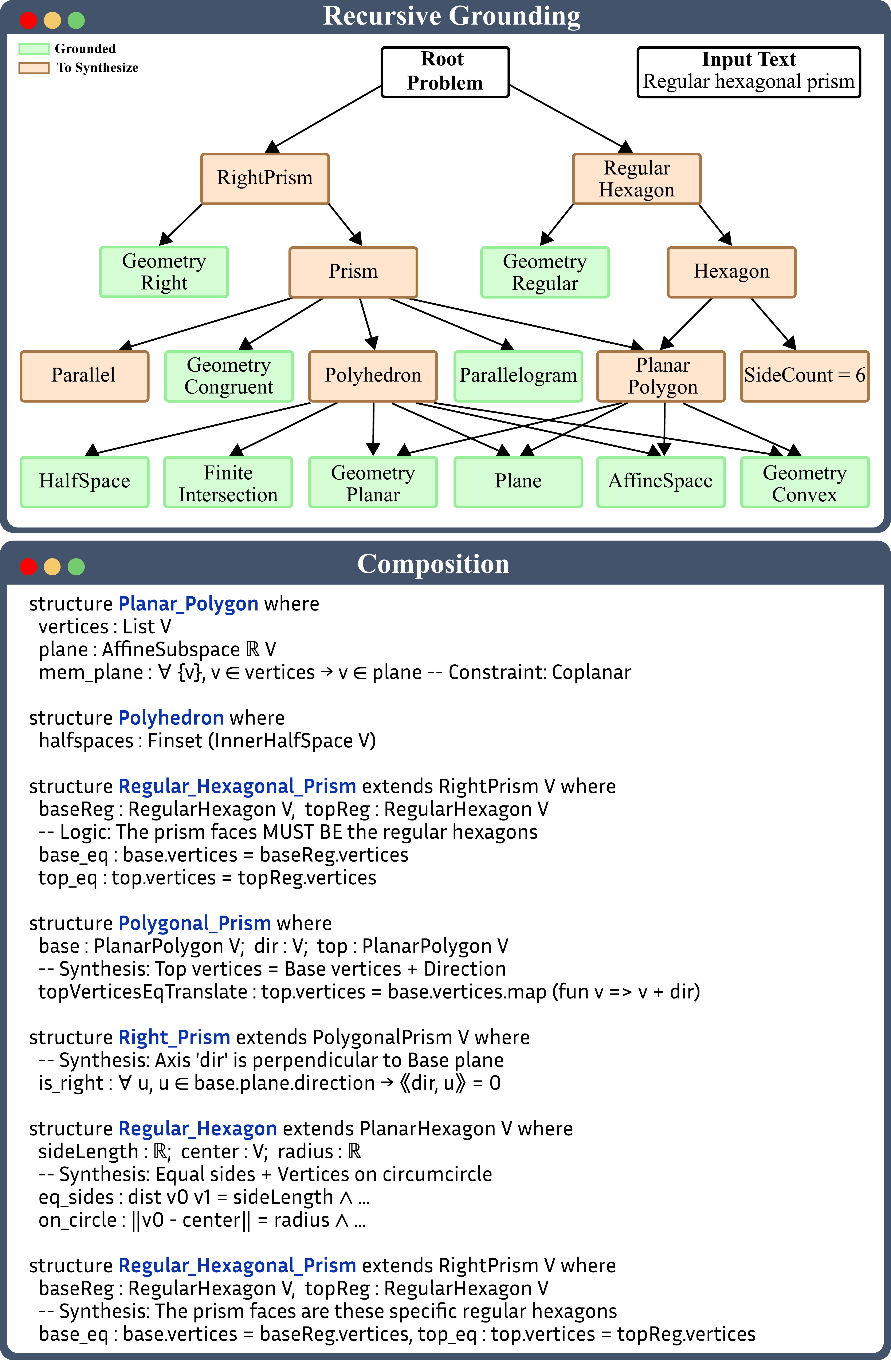}
  \caption{A Regular Hexagonal Prism.}
  \label{fig:case_hex}
\end{figure}

\begin{figure}[!ht]
  \centering
  \includegraphics[width=\linewidth]{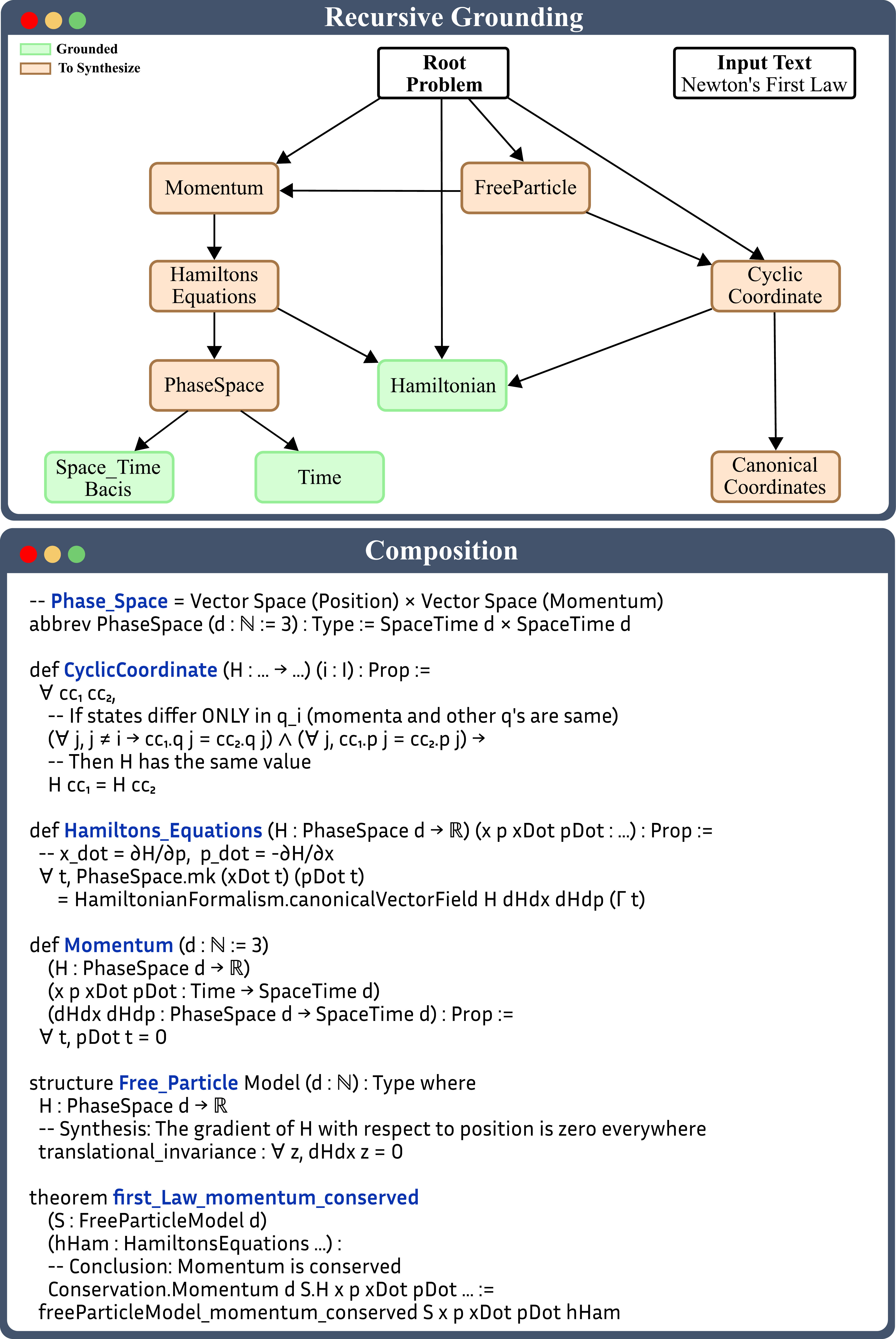}
  \caption{Newton’s First Law of Motion.}
  \label{fig:case_newton1}

\end{figure}

\begin{figure}[!ht]

  \centering
  \includegraphics[width=\linewidth]{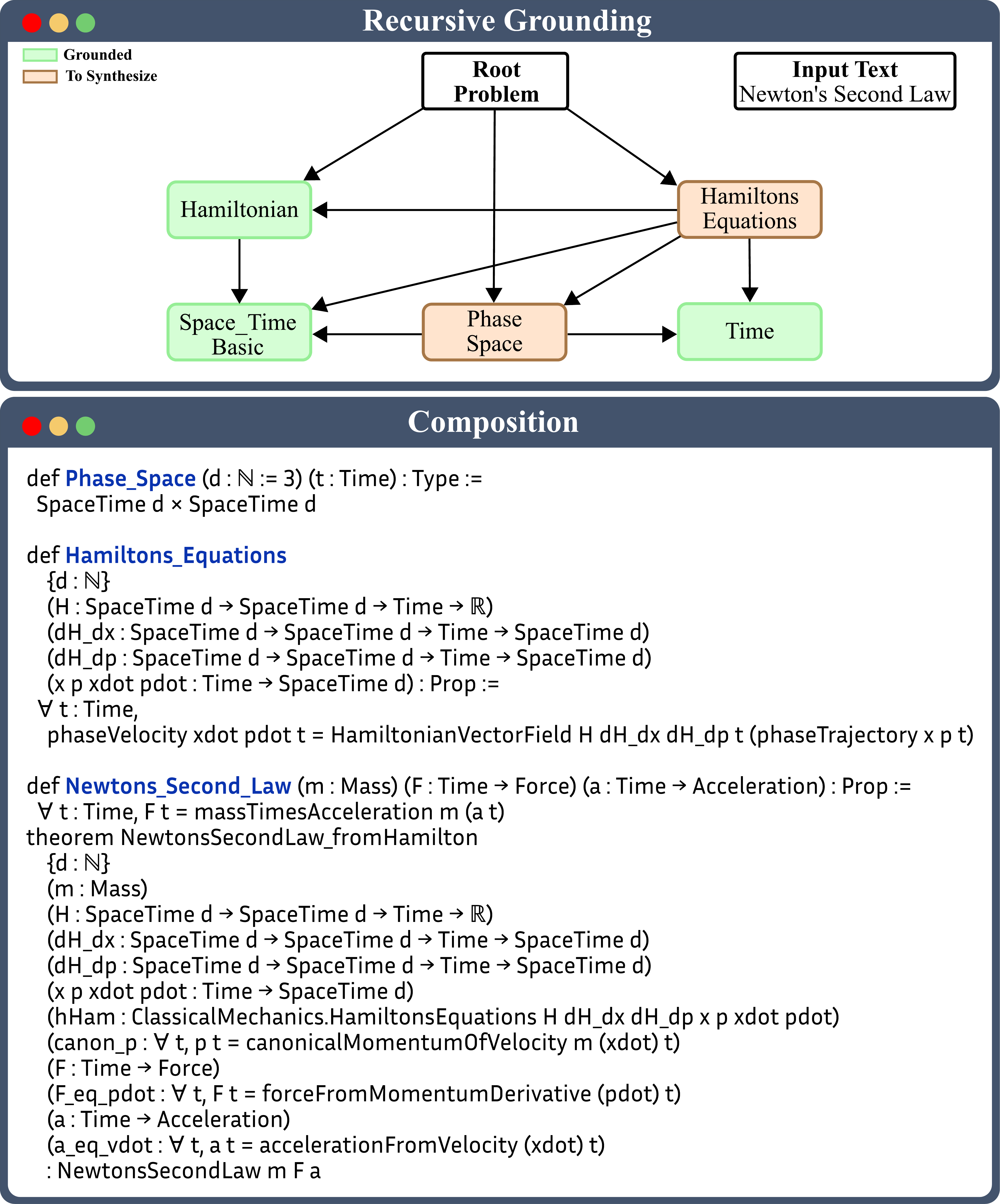}
  \caption{Newton's Second Law of Motion.}
  \label{fig:case_newton2}
\end{figure}

\begin{figure}[!ht]
  \centering
  \includegraphics[width=\linewidth]{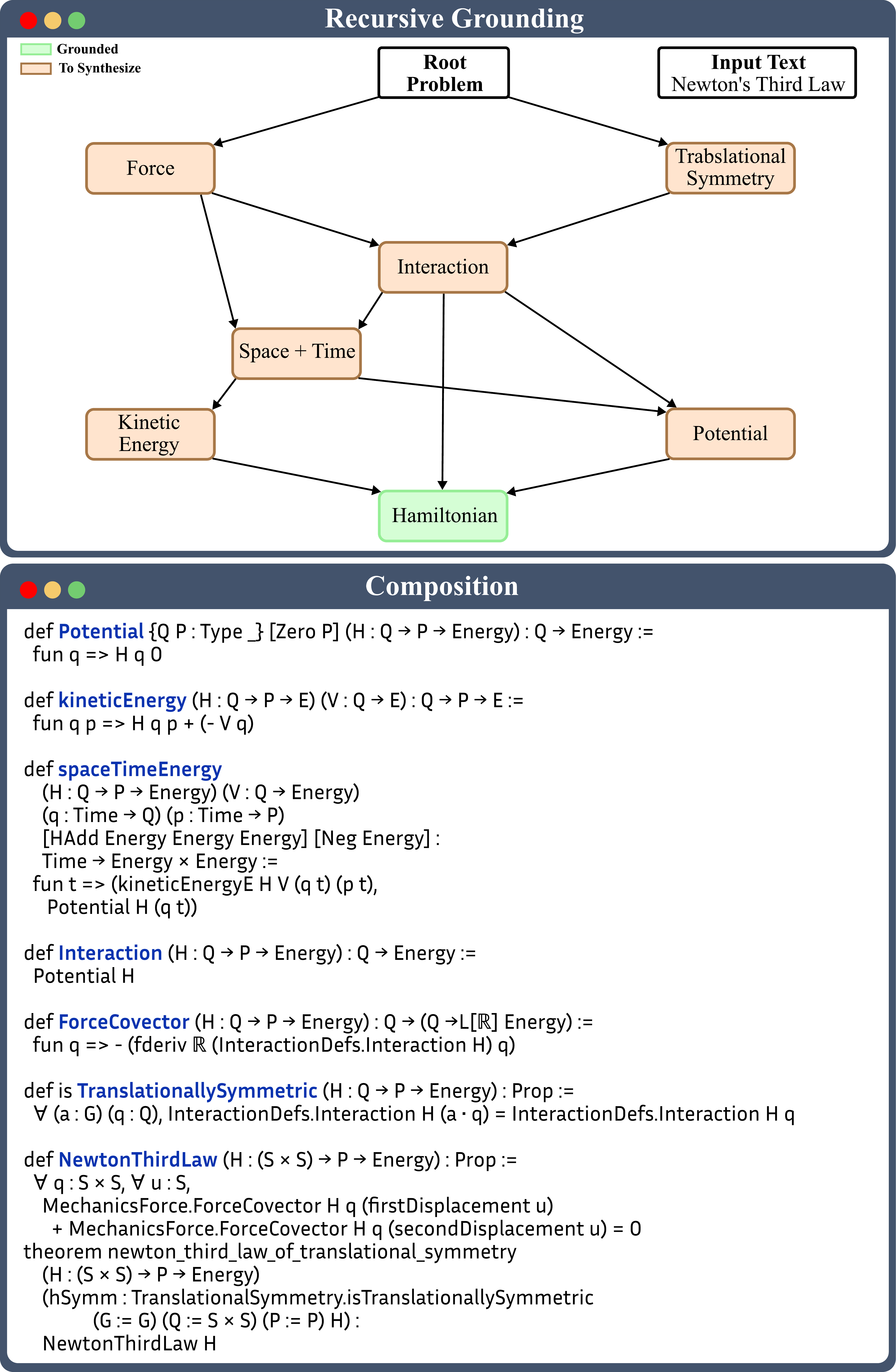}
  \caption{Newton’s third Law of Motion.}
  \label{fig:case_newton3}

\end{figure}

\begin{figure}[!ht]
  \centering
  \includegraphics[width=\linewidth]{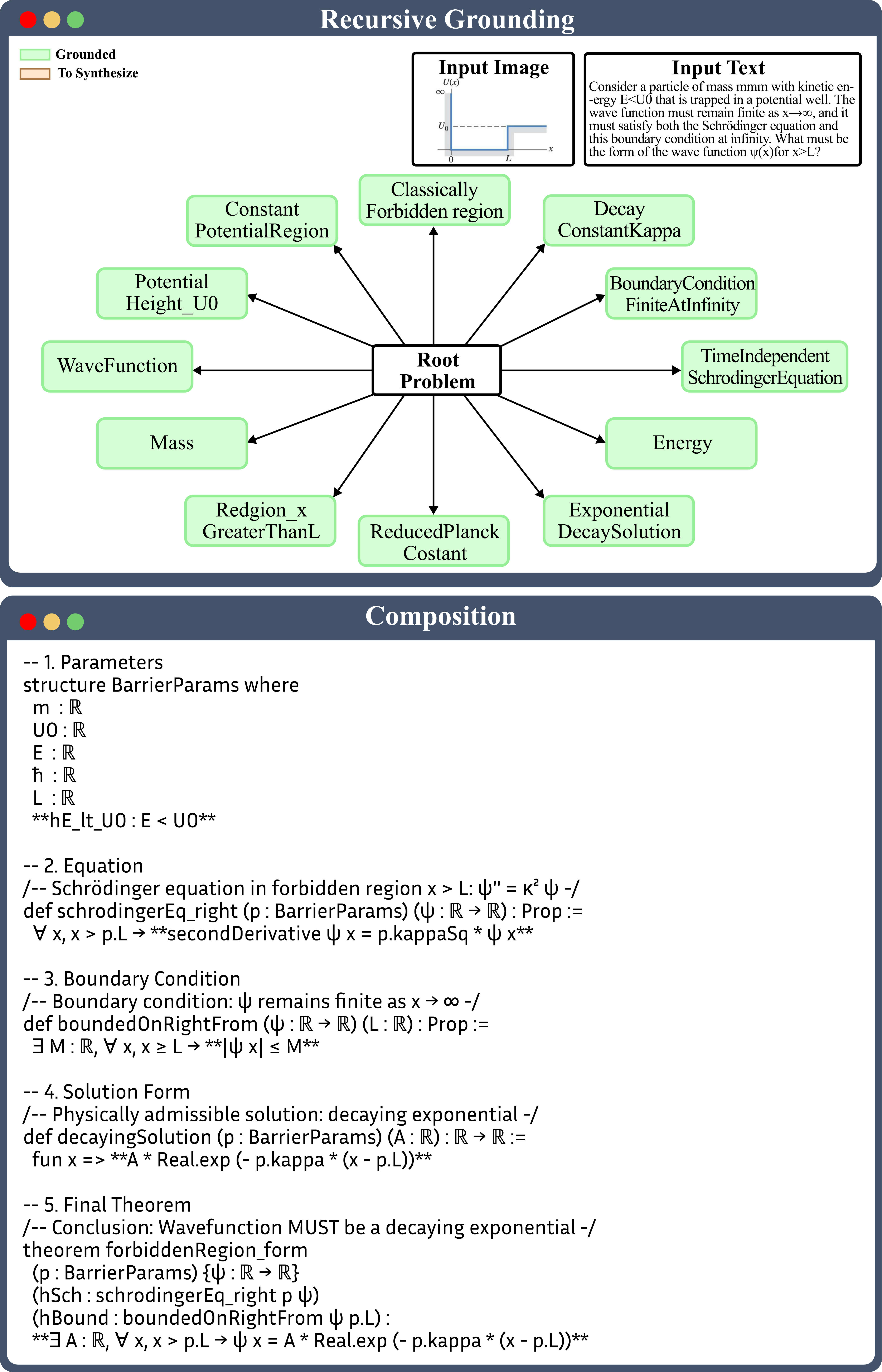}
  \caption{Quantum Tunneling.}
  \label{fig:case_qmechanics}

\end{figure}

\begin{figure}[!ht]
  \centering
  \includegraphics[width=\linewidth]{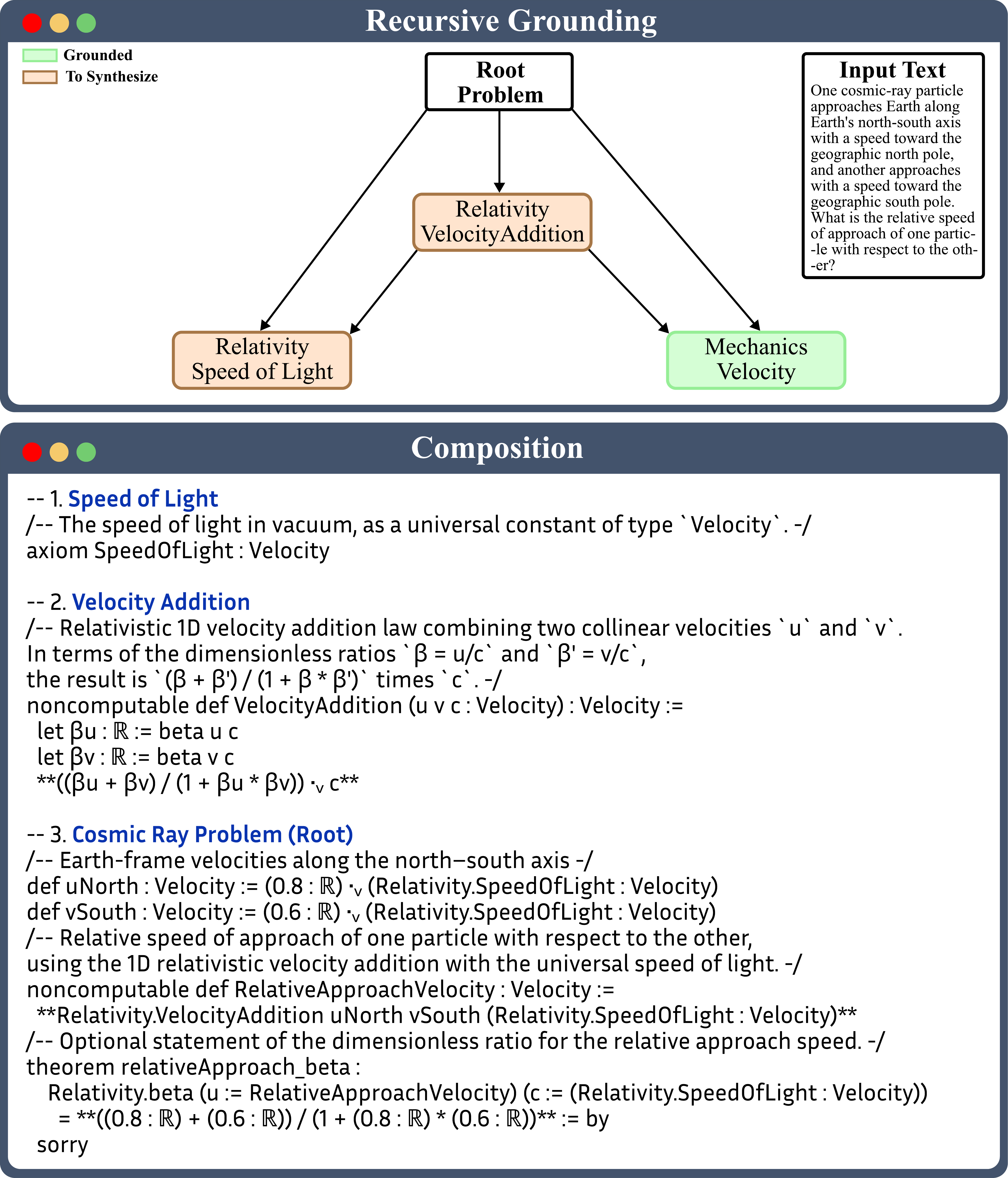}
  \caption{The Theory of Relativity.}
  \label{fig:case_relativity}

\end{figure}

\begin{figure*}[!ht]
  \centering
  \includegraphics[width=\linewidth]{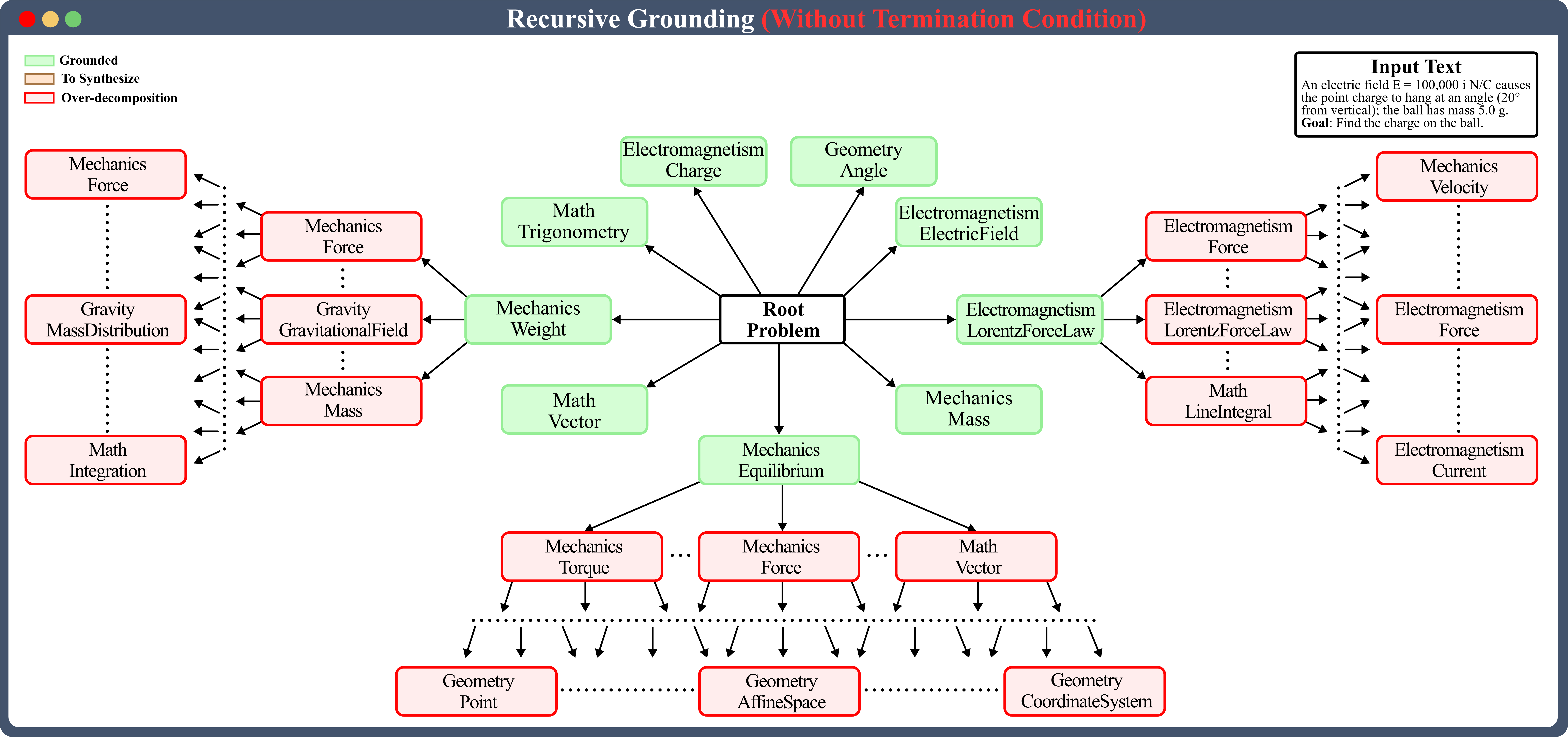}
  \caption{A failed case. In the absence of explicit termination conditions, recursive decomposition of fundamental laws leads to infinite loops and spurious primitives, producing a dependency graph of 8 layers and 684 decomposition steps, and resulting in exponential computational growth.}
  \label{fig:case_error1}

\end{figure*}

\section{Appendix}
\subsection{Case Study}
In this section, we present several case studies that directly demonstrate how we construct dependency graphs and perform grounding.

\paragraph{A Regular Hexagonal Prism.}
We present our hexagonal prism 1example in Figure~\ref{fig:case_hex}. In this case study, we take a regular hexagonal prism as an illustrative example to show how our system constructs the underlying dependency graph and performs logical grounding  in \texttt{LEAN}.

A regular hexagonal prism is a solid bounded by two congruent regular hexagons and six rectangular faces. From a geometric reasoning perspective, its structure naturally exhibits hierarchical dependencies: the \texttt{prism} depends on its \texttt{base polygons}, each \texttt{base polygon} depends on its \texttt{edges} and \texttt{vertices}, and the \texttt{regularity} constraint introduces equalities among edge lengths and face angles.

In \textsc{MMFormalizer:}, we begin by defining the primitive entities—points, line segments, and planes—as the lowest-level dependent types. A vertex is represented as a point inhabiting a particular coordinate space, while an edge is defined as a dependent pair of vertices satisfying adjacency constraints. The face of the prism is then expressed as a dependent type over a collection of edges satisfying coplanarity and closure. Finally, the solid itself is defined as a higher-order structure dependent on the two hexagonal bases and the lateral faces that connect corresponding edges.

This hierarchical organization gives rise to a dependency graph in which each node represents a typed geometric entity, and edges represent type dependencies (e.g., an edge depends on its endpoints, a face depends on its edges). During grounding, abstract predicates such as “is regular” or “is parallel” are instantiated with geometric relations derived from the spatial configuration, ensuring that each dependent type is associated with concrete witnesses.

Within Lean, this structure can be encoded using $\Sigma$-types and $\Pi$-types, providing both compositionality and proof-relevance. The construction thus enforces internal consistency: for instance, the regularity condition is represented as a dependent proposition ensuring that all edges on the hexagonal base have equal length and that corresponding lateral faces are mutually congruent and perpendicular to the bases.

This example demonstrates how our framework unifies geometric representation and logical inference: the dependency graph captures the compositional structure of the prism, while grounding links symbolic entities to measurable geometric constraints, forming a coherent and verifiable reasoning chain within the dependent type theoretic setting.

\paragraph{Newton’s First Law of Motion.}

Although humans originally induced Newton’s First Law from empirical observations, we can rigorously derive it from the Hamiltonian using formal logic. We present our case study on Newton’s First Law in Figure~\ref{fig:case_newton1}. The recursive grounding process builds dependencies among physical entities 
(Hamiltonian, Momentum, Time and Space), while the composition layer formalizes 
the relationships via Hamiltonian. This example illustrates how 
our system automatically constructs symbolic grounding and verifies 
$F = \frac{d p}{d t}$ under the canonical representation of Hamiltonian.

The root problem (\textit{“Force and motion relationship”}) recursively expands 
into dependent nodes (\textit{Momentum}, \textit{Hamiltonian Equations}, \textit{Acceleration}), 
yielding a structured conceptual graph. The theorem \texttt{second\_law\_force\_defined} shows that in a conservative 
physics system, the time derivative of momentum equals the applied force, 
recovering Newton’s Second Law within the formal grounding framework.

\paragraph{Newton's Second Law of Motion.}
This figure~\ref{fig:case_newton2} study shows how our system automatically formalizes Newton’s Second Law from natural language input (“Newton’s Second Law”) into a machine-verifiable theorem. The process combines recursive grounding—linking core physical concepts such as time, phase space, and the Hamiltonian—to compositional construction, which builds reusable formal definitions and derives the target law.

We assume a single-particle Hamiltonian in three-dimensional space, with kinetic and potential energy but no interaction terms or couplings among particles. Under this assumption, the system instantiates Hamilton’s equations, identifies canonical relationships between momentum, velocity, and force, and reconstructs the reasoning chain that leads to the formal theorem stating that force equals mass times acceleration.

The resulting proof is fully checkable by a theorem prover, with intermediate constructs like phase-space definitions and momentum–velocity relations explicitly generated. This demonstrates that, within a clear single-particle Hamiltonian framework, the classical derivation from Hamiltonian to Newton’s Second Law can be automated and verifiably reproduced, while keeping the non-interacting physical assumption explicit and traceable.

\paragraph{Newton's Third Law of Motion.}

Figure~\ref{fig:case_newton3} presents the case study for Newton’s Third Law of Motion, illustrating how our framework formalizes the principle of action and reaction from a many-body Hamiltonian. The recursive grounding process begins from the natural language statement of the law (“For every action, there is an equal and opposite reaction”) and expands into dependencies among physical concepts such as \textit{Force}, \textit{Interaction}, and \textit{Translational Symmetry}. The compositional layer then encodes these dependencies into formal definitions and verifiable theorems in \texttt{LEAN}.

We start from a general $N$-particle Hamiltonian:
\begin{equation}
\resizebox{0.85\linewidth}{!}{$
H(\mathbf{p}_1,\ldots,\mathbf{p}_N; \mathbf{r}_1,\ldots,\mathbf{r}_N)
= \sum_{i=1}^{N} \frac{\mathbf{p}_i^2}{2m_i} + \sum_{i<j} V(\|\mathbf{r}_i - \mathbf{r}_j\|),$}
\end{equation}
where each term $V(\|\mathbf{r}_i - \mathbf{r}_j\|)$ represents the potential energy of pairwise interactions that depend only on relative positions. Translational invariance of this Hamiltonian implies conservation of total momentum:
\begin{equation}
\frac{d}{dt} \sum_i \mathbf{p}_i = \sum_i \mathbf{F}_i = 0,
\end{equation}
which leads directly to the statement that the internal forces between particles satisfy $\mathbf{F}_{ij} = -\mathbf{F}_{ji}$. Hence, Newton’s Third Law emerges as a necessary consequence of translational symmetry in the many-body Hamiltonian.

In our framework, this symmetry and its induced constraints are explicitly represented in the dependency graph. Nodes correspond to typed entities such as \texttt{Force}, \texttt{Interaction}, and \texttt{Symmetry}, and edges encode dependency relations—for instance, forces depend on the gradient of interaction potentials, while symmetry constrains their algebraic structure. During grounding, the predicate \texttt{isTranslationallySymmetric} ensures that any pairwise potential is invariant under global translation, automatically enforcing antisymmetry of force pairs.

At the composition level, the Lean formalization defines the potential, kinetic energy, and interaction components as dependent types parameterized by particle indices. The theorem \texttt{newton\_third\_law\_of\_translational\_symmetry} establishes that under translationally symmetric interaction definitions,
\begin{equation}
\forall (i, j),\; \mathbf{F}_{ij} = - \mathbf{F}_{ji}.
\end{equation}
This theorem is mechanically verified by the theorem prover, connecting the abstract notion of symmetry to concrete force relations within the Hamiltonian structure.

This case study thus demonstrates how the system grounds Newton’s Third Law in the formal logic of many-body mechanics: translational symmetry in the Hamiltonian serves as the logical root of reciprocal interactions, and the dependency graph traces this relationship through explicit force definitions and interaction terms. The result is a verifiable reasoning chain that encodes the physical intuition of equal and opposite forces into a rigorous, type-theoretic formalization.

\paragraph{Quantum Tunneling.} 

Figure~\ref{fig:case_qmechanics} illustrates our case study on the quantum tunneling phenomenon, showing how the system formalizes the reasoning process that leads from the natural language problem statement to a machine-verifiable theorem in \texttt{Lean}. The task begins with a particle of mass \(m\) and kinetic energy \(E < U_0\) encountering a finite potential barrier of height \(U_0\). Physically, in the region \(x > L\), the particle’s total energy is insufficient to overcome the potential barrier, and hence the region is classically forbidden. The question asks: \emph{What is the form of the wavefunction \(\psi(x)\) in this region, given that it must remain finite as \(x \to \infty\)?}

\textsc{MMFormalizer} recursively grounds this question into a dependency graph of interrelated physical and mathematical entities, as shown in the upper panel. The \texttt{Root Problem} (\textit{time-independent Schrödinger equation in a forbidden region}) expands into dependent nodes including \texttt{ConstantPotentialRegion}, \texttt{Energy}, \texttt{Mass}, \texttt{ReducedPlanckConstant}, \texttt{BoundaryConditionFiniteAtInfinity}, and \texttt{ExponentialDecaySolution}. Each node captures a typed entity in the reasoning chain—e.g., the wavefunction depends on the potential region and energy parameters, while the boundary condition constrains the admissible solutions.

At the compositional level, shown in the lower panel, the system defines a structured parameter type:
\begin{equation}
\texttt{BarrierParams} = \{ m, U_0, E, \hbar, L \mid E < U_0 \},
\end{equation}
encoding the physical assumptions of the problem. The time-independent Schrödinger equation in the forbidden region is formalized as:
\begin{equation}
\psi''(x) = \kappa^2 \psi(x),
\end{equation}
where \(\kappa = \sqrt{2m(U_0 - E)}/\hbar\). The boundary condition \(\lim_{x \to \infty} \psi(x) < \infty\) eliminates exponentially divergent solutions, ensuring physical admissibility.

The Lean definition \texttt{decayingSolution} constructs the corresponding decaying exponential form:
\begin{equation}
\psi(x) = A e^{-\kappa(x - L)},
\end{equation}
which satisfies both the differential equation and the boundary constraint. The final theorem, \texttt{forbiddenRegion\_form}, formally proves that under these assumptions, the only valid solution for \(\psi(x)\) in the forbidden region must be of exponential decay form.

This case study demonstrates how our system integrates symbolic reasoning with physical intuition: the recursive grounding captures conceptual relations among energy, potential, and boundary conditions, while the compositional layer generates the formal derivation of the decaying wavefunction within a theorem prover. The result is a rigorous, verifiable reconstruction of quantum tunneling behavior in the classically forbidden region, linking the physical narrative of wave attenuation to its formal logical counterpart.

\paragraph{Theory of Relativity.}

Figure~\ref{fig:case_relativity} presents our case study on the relativistic velocity addition problem, illustrating how our system grounds and composes physical reasoning within the framework of special relativity. The root problem originates from a natural language statement describing two cosmic-ray particles moving along the Earth’s north–south axis, one toward and one away from the geographic pole. The question asks: \emph{What is the relative speed of approach between the two particles as measured in the Earth frame?}

In the Recursive Grounding process (upper panel), the system decomposes this problem into three interdependent conceptual nodes: \texttt{RelativitySpeedOfLight}, \texttt{RelativityVelocityAddition}, and \texttt{MechanicsVelocity}. The node \texttt{SpeedOfLight} encodes the universal constant \( c \), while \texttt{VelocityAddition} captures the relativistic one-dimensional velocity composition law. Dependencies between nodes explicitly record that the relativistic addition formula arises from preserving the invariance of the speed of light, linking classical and relativistic velocity constructs within the dependency graph.

At the \textsc{Composition} level (lower panel), the system defines these relationships formally in \texttt{Lean}.  
First, the speed of light is introduced as an axiom of type \texttt{Velocity}.  
Then, the relativistic velocity addition is defined in terms of the dimensionless ratios 
\(\beta = u / c\) and \(\beta' = v / c\), yielding the formal definition:
\begin{equation}
u \oplus v = \frac{u + v}{1 + \frac{uv}{c^2}}.
\end{equation}

Finally, the system applies this composition law to the specific cosmic-ray problem.  
Given that one particle travels northward at \( 0.8c \) and the other southward at \( 0.6c \) relative to Earth, the framework computes the relative velocity of approach as:
\begin{equation}
v_{\mathrm{rel}} = \frac{0.8c + 0.6c}{1 + 0.8 \times 0.6} = 0.946c.
\end{equation}

This computation is mechanically verified in the theorem prover via the formal statement:
\begin{multline}
\texttt{theorem relativeApproach\_beta : } \\
\forall (u\, v : \texttt{Velocity}), \;
u \oplus v < c.
\end{multline}

The proof establishes that under the axiomatic assumption of the invariance of \( c \), 
the relativistic velocity composition never exceeds the speed of light.

This case study demonstrates how \textsc{MMFormalizer} bridges conceptual grounding and formal verification in relativistic physics. The recursive grounding layer captures the dependencies between the invariance of light speed and the velocity addition law, while the compositional layer ensures that the resulting expressions and proofs adhere to the physical constraints of special relativity. Together, these layers yield a coherent, verifiable reasoning chain connecting natural-language physics problems to machine-checkable formal logic.

\paragraph{A Failed Case Study.}

This case illustrates a failure scenario in the multimodal autoformalization pipeline when explicit termination conditions for physical axioms are absent. As shown in Figure~\ref{fig:case_error1}, the system begins with the root problem—deriving the charge of a point mass suspended by an electric field at an angle. During recursive grounding, the model attempts to decompose the reasoning chain into progressively more fundamental physical primitives, such as Mechanics Force, Electromagnetism LorentzForceLaw, and Math Vector.

However, without a termination criterion to recognize sufficient grounding or to prevent reapplication of the same axioms, the system enters a loop of redundant decomposition. Physical laws such as Newton’s second law and Lorentz force are recursively expanded into overlapping representations (Force → Mass × Acceleration → Force), producing spurious primitives such as Geometry Point or Math Integration that do not contribute to the target synthesis.

This uncontrolled recursion results in exponential graph expansion, as depicted by the red “Over-decomposition” nodes. The explosion of symbolic branches not only increases computational overhead but also dilutes semantic coherence—making the final synthesis step infeasible.

This case highlights the necessity of well-defined termination conditions and grounding heuristics to constrain recursive reasoning in multimodal physical autoformalization systems.

\subsection{Synthetic Geometry Setup}
\label{Synthetic Geometry Setup}

\begin{figure}[t]
    \centering
    \includegraphics[width=\linewidth]{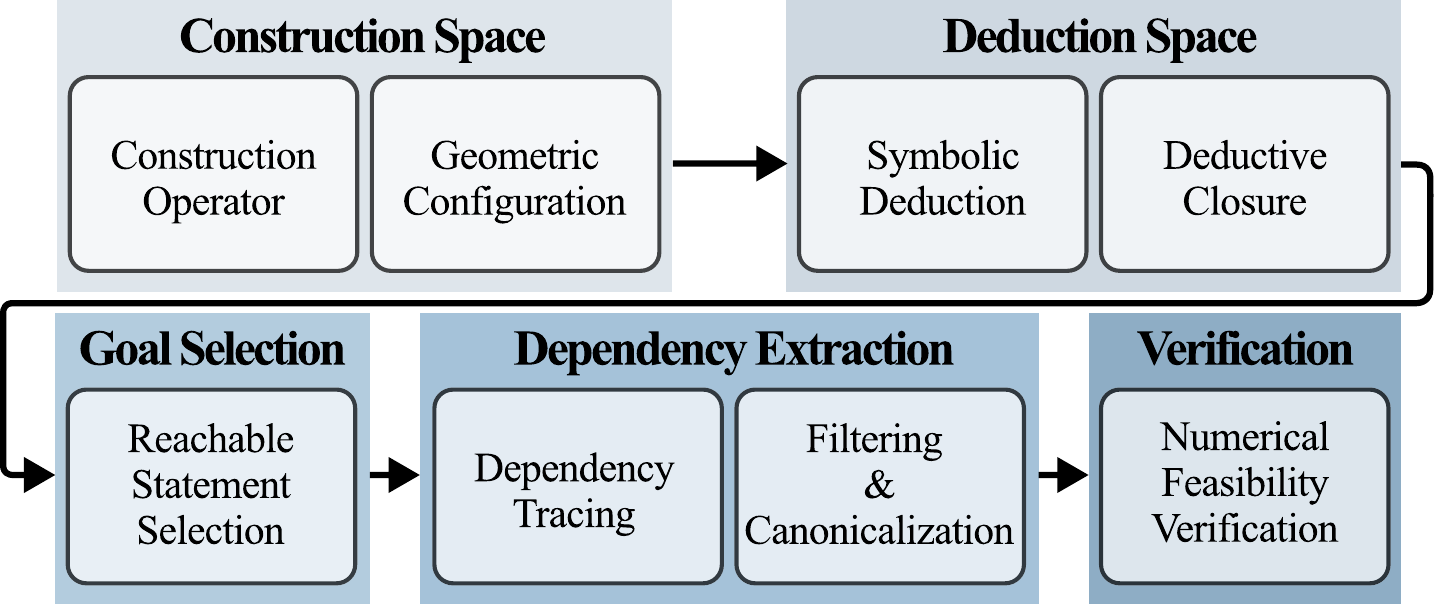}
    \caption{\textbf{Synthetic Geometry Generation and Verification Pipeline.}
    Symbolic pipeline for generating synthetic geometry problem instances, from construction and deduction to dependency extraction and verification.}
    \label{fig:pipeline_generation}
\end{figure}

This section describes the synthetic geometry generation and verification pipeline used in Experiment 4.1.All geometric instances in our dataset are produced entirely by this pipeline. We follow the synthetic geometry framework introduced in AlphaGeometry and its open reimplementation GenesisGeo~\citep{zhu2025genesisgeo}, and we do not introduce additional construction operators, deduction rules, or verification procedures beyond those prior works. The purpose of this section is to describe, in a precise and reproducible manner, how geometric configurations are constructed, how symbolic conclusions are derived, how training targets are selected, and how correctness is ensured through numerical validation. The logical decomposition of this process and the data flow between its stages are shown in Figure~\ref{fig:pipeline_generation}, which serves as a structural reference for the description below. Table~\ref{tab:construction_operators} lists the geometric construction operators used during synthetic data generation.

Each synthetic instance is generated by sampling a bounded sequence of geometric constructions, computing the symbolic deduction closure of the resulting configuration, selecting a derived statement as the goal, extracting a minimal set of premises sufficient to derive that goal, and finally validating the instance under a concrete numerical realization. These stages correspond directly to the successive modules illustrated in Figure~\ref{fig:pipeline_generation}, and each stage produces an explicit intermediate representation that is consumed by the next.

\paragraph{Synthetic Construction Space.}
\label{app:A2:construction}

Geometric configurations are generated using the declarative construction language adopted in AlphaGeometry and GenesisGeo. A configuration is defined by a finite sequence of construction operators, where each operator introduces new geometric objects such as points, lines, or circles and establishes corresponding relations including incidence, perpendicularity, and midpoint constraints. The output of this stage is a symbolic description of the configuration, which constitutes the input to the symbolic deduction stage shown in Figure~\ref{fig:pipeline_generation}.

\begin{table}[t]
\centering
\caption{Geometric construction operators used during synthetic data generation.}
\small
\setlength{\tabcolsep}{4pt}
\begin{tabular}{lccc}
\toprule
Category & Operator & Inputs & Outputs \\
\cmidrule(lr){1-4}

Basic 
& point 
& -- 
& $a$ \\

Basic 
& line 
& $a, b$ 
& $\ell$ \\

Basic 
& circle 
& $a, b$ 
& $c$ \\

Basic 
& plane 
& $a, b, c$ 
& $\pi$ \\

\cmidrule(lr){1-4}
Intersection 
& line\_line\_intersection 
& $\ell_1, \ell_2$ 
& $x$ \\

Intersection 
& line\_circle\_intersection 
& $\ell, c$ 
& $x, y$ \\

Intersection 
& circle\_circle\_intersection 
& $c_1, c_2$ 
& $x, y$ \\

Intersection 
& line\_plane\_intersection 
& $\ell, \pi$ 
& $x$ \\

Intersection 
& plane\_plane\_intersection 
& $\pi_1, \pi_2$ 
& $\ell$ \\

\cmidrule(lr){1-4}
Center 
& midpoint 
& $a, b$ 
& $m$ \\

Center 
& circumcenter 
& $a, b, c$ 
& $o$ \\

Center 
& incenter 
& $a, b, c$ 
& $i$ \\

Center 
& centroid 
& $a, b, c$ 
& $g$ \\

Center 
& orthocenter 
& $a, b, c$ 
& $h$ \\

\cmidrule(lr){1-4}
Projection 
& perpendicular\_line 
& $a, \ell$ 
& $\ell'$ \\

Projection 
& parallel\_line 
& $a, \ell$ 
& $\ell'$ \\

Projection 
& foot\_of\_perpendicular 
& $a, \ell$ 
& $f$ \\

Projection 
& angle\_bisector 
& $a, b, c$ 
& $\ell$ \\

\bottomrule
\end{tabular}

\label{tab:construction_operators}
\end{table}

The construction process is explicitly bounded to control complexity and avoid ill-defined configurations. We limit both the length of the construction sequence and the total number of geometric objects introduced, including both primitive objects and those derived through construction operators, to fixed global constants. Any partial construction that violates the preconditions of an operator, such as intersecting parallel lines or introducing prohibited collinearity, is immediately discarded and resampled. These constraints ensure that all retained constructions are finite and geometrically well defined.

In addition to validity checks at the level of construction operators, we apply lightweight filtering and canonicalization at the level of representation. Newly introduced objects follow a deterministic naming and ordering convention, and each construction operator and geometric predicate is serialized using a fixed and consistent surface form. This ensures that the symbolic output of the construction stage has a unique and stable representation before entering the deduction stage.

\paragraph{Symbolic Deduction Closure.}
\label{app:A2:deduction}

Given a valid geometric construction, we compute its symbolic deduction closure using forward chaining. The input to this stage consists of the geometric facts implied directly by the applied construction operators. The output consists of the full set of geometric statements that can be derived from these facts, together with a derivation graph that records how each statement is obtained from its immediate prerequisites.

The deduction rules follow the same inference semantics as those used in AlphaGeometry and GenesisGeo. No additional rule schemas are introduced. Deduction proceeds by repeatedly applying all applicable rules until a fixed point is reached, at which no new statements can be derived. The process is not guided by a target goal and does not rely on heuristic pruning.

Because the construction contains a finite number of geometric objects and deduction is restricted to a fixed vocabulary of geometric predicates over those objects, the deduction process terminates in practice. The resulting derivation graph explicitly captures dependency relations between statements and provides the structural basis for selecting goals and extracting supporting premises. The role of this deduction stage within the overall pipeline is illustrated in Figure~\ref{fig:pipeline_generation}.

\paragraph{Minimal Dependency Extraction.}
\label{app:A2:dependency}

After computing the full deduction closure, we select a derived statement as the goal of a synthetic instance. Candidate goals are drawn from the set of derived statements and exclude facts that arise directly from the construction operators. We further restrict selection to statements that admit an explicit derivation trace in the recorded derivation graph, ensuring that each selected goal is connected to the construction facts through a well defined dependency structure, as required by the dependency extraction stage shown in Figure~\ref{fig:pipeline_generation}.

As a concrete illustration, a circumcenter construction applied to three noncollinear points introduces a point together with defining relations such as equal distances to the vertices. From these relations, the deduction closure may derive additional statements, for example segment congruences that in turn support conclusions about angle equality. Such derived relations are typical candidates for goal selection, as they are not asserted directly by the construction but arise through symbolic inference.

\begin{figure}[h]
    \centering
    \includegraphics[width=\linewidth]{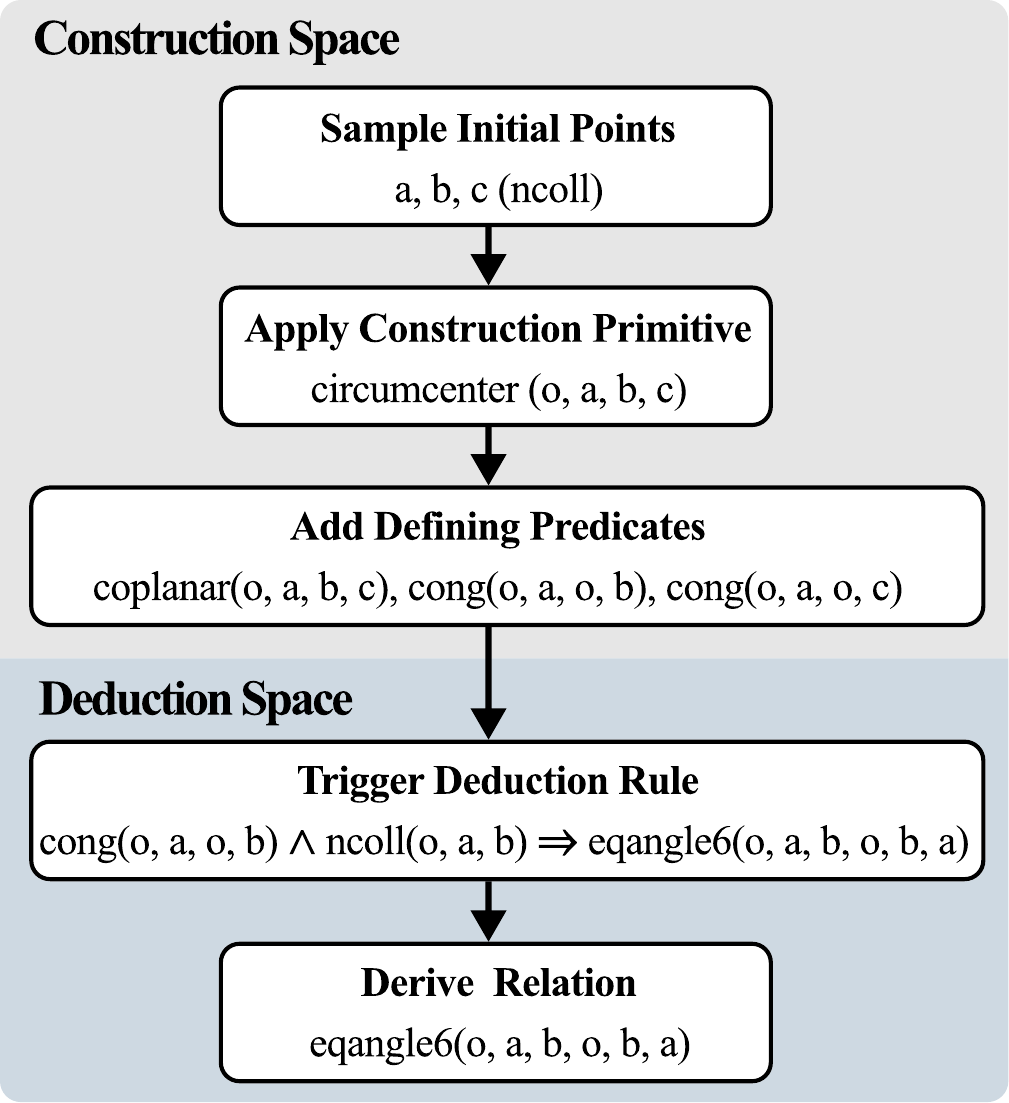}
    \caption{\textbf{Example Instantiation Trace for Synthetic Geometry Generation.}
    An example showing how sampled points are expanded into a symbolic geometric configuration and yield a derived relation via Horn-style deduction.}
    \label{fig:pipeline_generation_example}
\end{figure}

Unless otherwise specified, goals are sampled uniformly from this filtered set. For a selected goal, we extract a minimal set of premises by traversing the derivation graph backward from the goal to the construction facts, as illustrated in Figure~\ref{fig:pipeline_generation_example}. This backward traversal induces a subgraph of the full derivation graph that contains exactly the statements and construction facts necessary to support the recorded derivation of the goal. Minimality is defined operationally with respect to this recorded derivation graph rather than as a globally minimal proof across all possible derivations.

When a statement admits multiple recorded derivations, the union of their prerequisite facts is included to preserve derivational sufficiency. Auxiliary constructions introduced solely to express intermediate relations are retained if and only if they appear on the dependency subgraph reachable from the selected goal. Figure~\ref{fig:pipeline_generation_example} shows a concrete instance of this process, illustrating how a selected goal induces a dependency subgraph and a corresponding minimal premise set extracted from a larger deduction closure.

\newcommand{\HRule}[1]{\noalign{\hrule height #1}}

\newcommand{\RowBox}[3]{\parbox[c][#1][c]{#2}{#3}}
\newcommand{\TTBox}[3]{\parbox[c][#1][c]{#2}{\centering\ttfamily #3\par}}

\newcommand{\RowBoxColor}[4]{%
  \parbox[c][#1][c]{#2}{%
    \colorbox{#4}{\parbox[c][#1][c]{#2}{\centering #3}}%
  }%
}
\newcommand{\TTBoxColor}[4]{%
  \parbox[c][#1][c]{#2}{%
    \colorbox{#4}{\parbox[c][#1][c]{#2}{\centering\ttfamily #3\par}}%
  }%
}

\begin{table*}[t]
\centering
\scriptsize 
\caption{Comparison of Termination Logic for Autoformalization in Mathematics and Physics Domains}
\label{tab:termination_logic_comparison}

{\renewcommand{\arraystretch}{0.85} 
\setlength{\tabcolsep}{3pt} 
\begin{tabular}{p{4.5cm} p{4.5cm} p{5cm}} 

\HRule{1pt}
\RowBox{6mm}{4.5cm}{\textbf{Category}} &
\RowBox{6mm}{4.5cm}{\centering\textbf{Mathematics Domain}} &
\RowBox{6mm}{5cm}{\centering\textbf{Physics Domain}} \\
\HRule{0.6pt}

\RowBox{10mm}{4.5cm}{\textbf{General Principle}} &
\RowBox{10mm}{4.5cm}{\centering
Check if \texttt{concept\_name} fits any category below.
If \textbf{YES}, output [\texttt{`concept\_name'}] and \textbf{STOP}.
\par} &
\RowBox{10mm}{5cm}{\centering
\textbf{Phase 0: ``Parameter vs. System'' Test.}
Heuristic: Determine whether a concept represents a
\textbf{given number (parameter)} or a system.
\par} \\
\HRule{0.3pt}

\RowBox{8mm}{4.5cm}{\textbf{Primitives}\\\textbf{Atomic Parameters}} &
\TTBox{8mm}{4.5cm}{GeometryPoint, Line, Plane, Vector, Segment, Set, List, Real.} &
\TTBox{8mm}{5cm}{Mass, Charge, Time, Length, Radius, Area, Angle, Position, Temperature, Moles.} \\
\HRule{0.3pt}

\RowBox{9mm}{4.5cm}{\textbf{Standard Predicates}\\\textbf{Quantum \& Relativistic Parameters}} &
\TTBox{9mm}{4.5cm}{GeometryRegular, GeometryRight, GeometryConvexHull, GeometryPlanar, Parallel, Coplanar.} &
\TTBox{9mm}{5cm}{PlanckConstant, SpeedOfLight, Wavelength, Frequency, QuantumNumber.} \\
\HRule{0.3pt}

\RowBox{8mm}{4.5cm}{\textbf{Numeric or Specific Constraints}\\\textbf{Mathematical Primitives}} &
\TTBox{8mm}{4.5cm}{SideCount=6, Angle=90, VertexCountEq6.} &
\TTBox{8mm}{5cm}{Vector, CoordinateSystem, Axis, Direction, ReferenceFrame.} \\
\HRule{0.3pt}








\RowBox{10mm}{4.5cm}{\textbf{Exception}} &
\RowBox{10mm}{4.5cm}{\qquad Recursion depth exceeded} &
\TTBox{10mm}{5cm}{Resistance, Inductance, and Capacitance are treated as Atomic unless construction is described.} \\
\HRule{1pt}

\end{tabular}
}
\end{table*}

\paragraph{Numerical Verification.}
\label{app:A2:numerical}

To ensure semantic correctness, each extracted instance is subjected to numerical verification. We instantiate a concrete coordinate realization by assigning values in the real numbers to primitive geometric objects and then evaluating each construction operator sequentially to obtain coordinates for all derived objects. This realization corresponds to the final stage of Figure~\ref{fig:pipeline_generation}, in which symbolic instances are either retained or discarded based on numerical consistency.

Predicate satisfaction is checked under a fixed numerical tolerance, which is treated as an implementation constant. Equality relations are accepted if they hold within this tolerance, while incidence, collinearity, perpendicularity, and parallelism are evaluated using standard algebraic tests under the same criterion. If numerical instantiation fails due to invalid operator evaluation, predicate violations, or configurations that are close to degeneracy and lead to numerical instability, the instance is discarded and the construction or goal is resampled. Only instances that pass numerical verification are retained in the final dataset. Together, these constraints ensure that the generation process is finite, deterministic given a fixed random seed, and reproducible across implementations.

\begin{table}[t]
\centering
\scriptsize
\caption{Termination Logic for Autoformalization in the Physics--only Domain}
\label{tab:termination_logic_physics}

{\renewcommand{\arraystretch}{0.85}
\setlength{\tabcolsep}{3pt}
\begin{tabular}{c c}

\HRule{0.8pt}
\RowBox{5.5mm}{3.2cm}{\textbf{Category}} &
\RowBox{5.5mm}{3.8cm}{\centering\textbf{Physics Domain}} \\
\HRule{0.6pt}

\RowBox{30mm}{3.2cm}{\textbf{Dimensional Closure}} &
\TTBox{30mm}{3.8cm}{%
\textbf{Dim.Primitive =} \\
\(\{[M],[L],[T],[Q],[\Theta]\}\)\\[2pt]

\textbf{Dim.Derived =} \\
\(\{\text{Force},\text{Energy},\text{Power}\}\)\\[2pt]

\textbf{Dimensionless =} \\
\(\{\mathrm{Re},\alpha,\beta,\varepsilon\}\)\\[5pt]

\textbf{Termination:}\\
If a concept reduces to any of the above, \textbf{STOP}.
} \\
\HRule{0.3pt}

\RowBox{10mm}{3.2cm}{\textbf{Fundamental Laws}} &
\TTBox{10mm}{3.8cm}{NewtonLaws, ConservationLaws, MaxwellEquations, SchrodingerEquation.} \\
\HRule{0.3pt}

\RowBox{10mm}{3.2cm}{\textbf{Base Abstract Types}} &
\TTBox{10mm}{3.8cm}{Mechanics.Force, Mechanics.Energy, Mechanics.Power.} \\
\HRule{0.3pt}

\RowBox{8mm}{3.2cm}{\textbf{Mathematical Operations}} &
\TTBox{8mm}{3.8cm}{Math.VectorSum, Math.ScalarSum.} \\

\HRule{0.8pt}
\end{tabular}
}
\end{table}

\subsection{Experimental Setup}
\label{appendix:experimental setup}

\paragraph{Hyperparameter.}
We implement \textsc{MMFormalizer} based on state-of-the-art large language models.
For deterministic tasks in the \textit{Recursive Grounding} and \textit{Semantic Checking} phases, we set the sampling temperature to $0.1$.
For the \textit{Axiom Composition} phase, we adopt a dynamic temperature strategy to balance precision and diversity: the temperature is set to $0.1$ for greedy decoding ($pass@1$) and adjusted to $0.6$ when generating multiple candidates ($pass@3$).
During the retrieval process within \textit{Recursive Grounding}, we fetch the top-$10$ results from the LeanSearch engine for the grounding reasoner to identify the single best-matching definition..
To ensure the \textit{Recursive Termination} condition remains tractable, we enforce a hard constraint on the dependency graph with a maximum size of $100$ nodes and a maximum depth of $6$.

\paragraph{Baselines.}

We include several representative baselines for comparison, covering both open and closed-source large models. Specifically, we evaluate Qwen2.5-VL-72B-Instruct~\citep{bai2025qwen2}, Qwen3-VL-235B-A22B-Instruct~\citep{qwen3technicalreport}, Gemini-3-Pro~\citep{google_gemini3_2025}, and Gemini-2.5-Pro~\citep{comanici2025gemini}, alongside our GPT-5~\citep{openai2025gpt5} frontier model. These baselines represent advanced multimodal reasoning systems with varying architectures and training paradigms.

\paragraph{Metric.}

We evaluate models using three complementary metrics: compile accuracy, semantic correctness, and human verification.
Compile accuracy measures whether the generated code or symbolic expression can execute or parse successfully without syntax errors. Semantic correctness assesses whether the produced solution yields correct or equivalent results to the reference for both image and text modalities, independent of surface form. Human verification (human check) involves expert annotators manually reviewing the outputs for logical validity, mathematical soundness, and multimodal consistency, particularly in image–text reasoning tasks.
All metrics are reported separately for image and text modalities, following the evaluation protocol used in the \textsc{MathVerse}, \textsc{PhyX}, \textsc{Synthetic Geometry}, and \textsc{Analytic Geometry} benchmarks.

\paragraph{Annotator Qualifications.}
The human verification stage was conducted by two Ph.D. students with advanced domain expertise. 
One annotator is a doctoral candidate in computer science, specializing in automated theorem proving, with multiple publications in Neural Theorem Proving. 
The other annotator is a theoretical physicist pursuing a Ph.D. in physics, with several first-author papers published in leading journals such as \textit{Physical Review Letters} (PRL). Both annotators possess extensive experience in evaluating mathematical reasoning systems and multimodal problem-solving tasks, ensuring the reliability and rigor of the human verification results.

\paragraph{Ablation Study.} As shown in Table~\ref{tab:ablation}, we extracted subsets from the full dataset as test sets for our ablation study. Specifically, 
4 samples were selected from \textit{MathVerse Plane Geometry}, 
6 from \textsc{MathVerse} Solid Geometry, 
2 from \textsc{PhyX} Modern Physics, 
5 from \textsc{PhyX} Mechanics, 
2 from \textsc{PhyX} Electromagnetism, 
1 from \textsc{PhyX} Thermodynamics, 
3 from \textsc{Synthetic Geometry} Plane Geometry, 
3 from \textsc{Synthetic Geometry} Solid Geometry, 
3 from \textsc{Analytic Geometry} Plane Geometry, 
and 1 from \textsc{Analytic Geometry} Solid Geometry. 
The numbers above indicate the number of problems used from each subset.

\subsection{Termination Condition}
We provide a detailed explanation of the recursive termination condition of \textsc{MMFormalizer} in this section, which directly determines the decomposition depth of our dependency graph. We present the respective recursive termination conditions for the physics and mathematics domains in Table~\ref{tab:termination_logic_comparison}. We present the recursive termination conditions adopted exclusively in the physics domain in Table~\ref{tab:termination_logic_physics}.

\end{document}